\documentclass[a4paper,fleqn]{cas-sc}

\usepackage[authoryear]{natbib}

\usepackage{amssymb}
\usepackage{amsmath}

\usepackage{booktabs}
\usepackage{multirow}
\usepackage{caption}
\usepackage{tabularx} 
\usepackage{pifont}
\newcommand{\cmark}{\ding{51}}
\newcommand{\xmark}{\ding{55}}
\usepackage{xcolor}
\usepackage{makecell}
\usepackage{etoolbox}
\AtBeginEnvironment{tabular}{\rmfamily}
\AtBeginEnvironment{tabular*}{\rmfamily}
\usepackage{graphicx}
\usepackage{float}  
\usepackage{subcaption}   

\makeatletter
\ExplSyntaxOn
\cs_set_protected:Npn \__make_tbl_caption:nn #1#2
{
  \centering 
  \small
  \sffamily
  \textbf{#1.}~#2\par
  \vspace{10pt}
}
\ExplSyntaxOff
\makeatother

\def\tsc#1{\csdef{#1}{\textsc{\lowercase{#1}}\xspace}}
\tsc{WGM}
\tsc{QE}

\begin{document}
\let\WriteBookmarks\relax
\def\floatpagepagefraction{1}
\def\textpagefraction{.001}

\shorttitle{MedP-CLIP: Medical CLIP with Region-Aware Prompt Integration}    

\shortauthors{J. Peng et al.}  

\title [mode = title]{MedP-CLIP: Medical CLIP with Region-Aware Prompt Integration}  



\author[1]{Jiahui Peng}

\author[1]{He Yao}
\author[2]{Jingwen Li}
\author[3]{Yanzhou Su}
\author[4]{Sibo Ju}
\author[1]{Yujie Lu}
\author[5]{Jin Ye}
\author[6]{Hongchun Lu}
\author[2]{Xue Li}
\author[1]{Lincheng Jiang}

\author[1]{Min Zhu}
\cormark[1]

\author[1,5]{Junlong Cheng}[
                            orcid=0000-0002-6849-9093
]
\cormark[1]

\cortext[1]{Corresponding author}
\cortext[0]{Emails: zhumin@scu.edu.cn (M. Zhu); cjl951015@163.com (J. Cheng) }

\affiliation[1]{organization={Sichuan University},
            city={Chengdu},
            postcode={610065}, 
            country={China}}

\affiliation[2]{organization={Xinjiang University},
            city={Urumqi},
            postcode={830046}, 
            country={China}}

\affiliation[3]{organization={Alibaba Group},
            city={Hangzhou},
            postcode={310023}, 
            country={China}}

\affiliation[4]{organization={Fuzhou University},
            city={Fuzhou},
            postcode={350108}, 
            country={China}}

\affiliation[5]{organization={Shanghai AI Laboratory},
            city={Shanghai},
            postcode={200232}, 
            country={China}}

\affiliation[6]{organization={Southwest Jiaotong University},
            city={Chengdu},
            postcode={611756}, 
            country={China}}

\begin{abstract}
    Contrastive Language-Image Pre-training (CLIP) has demonstrated outstanding performance in global image understanding and zero-shot transfer through large-scale text-image alignment. However, the core of medical image analysis often lies in the fine-grained understanding of specific anatomical structures or lesion regions. Therefore, precisely comprehending region-of-interest (RoI) information provided by medical professionals or perception models becomes crucial. To address this need, we propose MedP-CLIP, a region-aware medical vision-language model (VLM). MedP-CLIP innovatively integrates medical prior knowledge and designs a feature-level region prompt integration mechanism, enabling it to flexibly respond to various prompt forms (e.g., points, bounding boxes, masks) while maintaining global contextual awareness when focusing on local regions. We pre-train the model on a meticulously constructed large-scale dataset (containing over 6.4 million medical images and 97.3 million region-level annotations), equipping it with cross-disease and cross-modality fine-grained spatial semantic understanding capabilities. Experiments demonstrate that MedP-CLIP significantly outperforms baseline methods in various medical tasks, including zero-shot recognition, interactive segmentation, and empowering multimodal large language models. This model provides a scalable, plug-and-play visual backbone for medical AI, combining holistic image understanding with precise regional analysis.
\end{abstract}




\begin{NoHyper}
\maketitle
\end{NoHyper}

\section{Introduction}

CLIP~\citep{CLIP} has achieved remarkable zero-shot generalization across diverse vision-language applications, including open-world recognition~\citep{ViLD,ODISE}, multimodal large language models (MLLMs)~\citep{BLIP-2,LLaVA}, and generative tasks in both 2D and 3D domains~\citep{BLIP-Diffusion,Point-E,PureCLIPNeRF}. This success stems from large-scale pretraining that captures holistic image-text alignment, enabling robust understanding of global semantics. However, CLIP and its derivatives struggle in domains requiring precise, fine-grained region-specific interpretation—most critically in medical imaging, where diagnostic accuracy hinges on localized anatomical structures or pathological lesions.

Medical image analysis fundamentally differs from natural image understanding. Clinical workflows are inherently localized: diagnoses often depend on identifying subtle, spatially confined anatomical structures or pathological regions, such as tumors, lesions, or organ boundaries. Existing CLIP adaptations for the medical domain, such as BiomedCLIP~\citep{zhang2025multimodal}, PMC-CLIP~\citep{lin2023pmc}, and MedCLIP~\citep{wang2022medclip}, enhance global semantic alignment by pretraining on domain-specific image-text corpora~\citep{lin2023pmc,wang2022medclip}. Yet, these models still lack native region-aware capabilities and remain global encoders, limiting their utility in clinical scenarios that require precise spatial reasoning.

In pursuit of region-aware understanding, several approaches have emerged in general computer vision. Naive strategies include isolating RoIs through cropping~\citep{ReCLIP,zhong2022regionclip} or suppressing background context via masking (applied at either the input or feature/attention level)~\citep{MaskCLIP,MaskAdaptedCLIP}. However, these methods suffer critical drawbacks: cropping inherently disrupts crucial global context, while masking neglects potentially informative surrounding structures and both undermine the holistic reasoning essential for accurate medical interpretation. Alternative frameworks like Alpha-CLIP~\citep{sun2024alpha} attempt to mitigate this by fusing RoI information via an auxiliary alpha channel input, preserving the full image context. Nevertheless, this architectural choice inherently ties the model's inference performance to the precision of the mask provided, exhibiting limited practicality in rapid clinical workflows due to cumbersome input requirements. Within the medical domain, methods such as R-LLaVA~\citep{chen2025r} and MedPLIB~\citep{huang2025towards} leverage bounding boxes or masks to guide large language models (LLMs) in contextualizing local regions. However, these solutions typically entail intricate, task-specific multimodal pipelines that incur substantial training costs and lack scalability across diverse clinical applications.

\begin{figure*}[pos=t] 
	\centering 
	\includegraphics[width=0.99\textwidth]{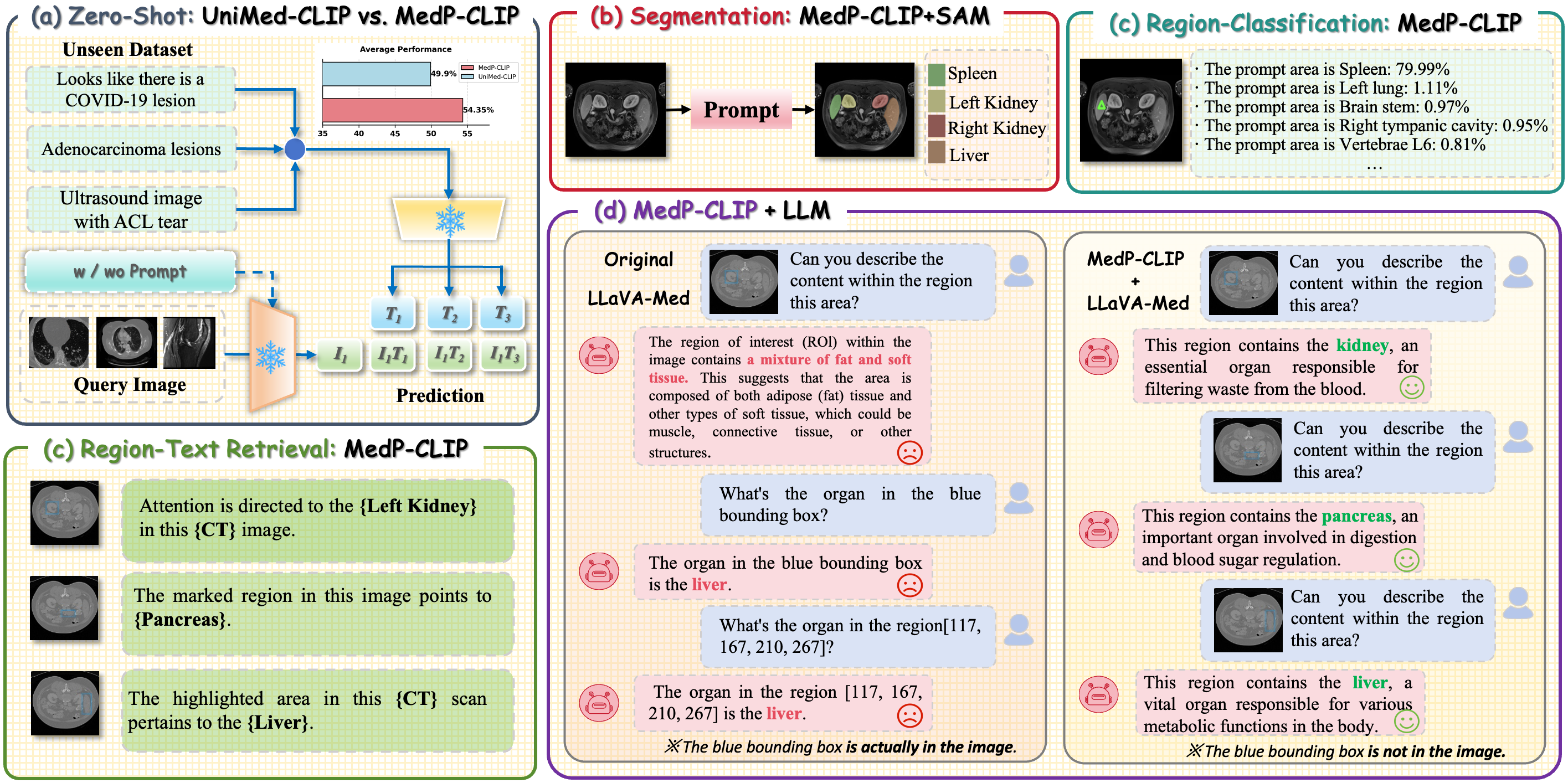} 
	\caption{MedP-CLIP can be flexibly deployed for various downstream tasks, significantly enhancing the region-aware analysis capability of medical imaging. This model supports direct application and plug-and-play integration, and accepts explicit prompts from users as well as outputs from perceptual models.}
	\label{fig1}
\end{figure*}

To overcome these limitations, we introduce MedP-CLIP, a region-aware VLM designed specifically for medical image analysis. Unlike approaches that compromise global context or impose impractical input requirements at inference, MedP-CLIP features a novel feature-level region prompt integration mechanism to achieve native promptability. While we utilize high-precision pixel-level masks during pre-training to establish accurate anatomical ground truth, our architecture is explicitly designed to decouple from them during deployment. By processing sparse embeddings through cross-attention, this design natively handles low-effort, arbitrary spatial prompts (including points, bounding boxes, and masks) while preserving comprehensive image context. Crucially, we synthesize 97.3 million regions by systematically integrating mask annotations from the IMed-361M~\citep{cheng2025interactive} dataset. This medical-specific optimization enables flexible, precise region understanding without task-specific pipelines or complex multimodal fusion, establishing a scalable backbone for diverse clinical applications.
We comprehensively evaluate the four core medical AI capabilities of MedP-CLIP (Fig.~\ref{fig1}), demonstrating its effectiveness in region-aware medical image understanding: \textbf{(1) Image Recognition (Fig.~\ref{fig1}(a)):} While preserving CLIP’s global recognition capabilities, MedP-CLIP uniquely enables prompting for arbitrary regions, improving average accuracy by 4.45\% over UniMed-CLIP on unseen datasets. \textbf{(2) Interactive Segmentation (Fig.~\ref{fig1}(b)):} As a plug-and-play backbone for interactive frameworks (e.g., SAM), MedP-CLIP achieves state-of-the-art segmentation performance with minimal fine-tuning costs ($\sim$1\% of total parameters). \textbf{(3) Region-Level Visual Grounding (Fig.~\ref{fig1}(c)):} By extending CLIP’s image-text alignment to regional semantics, our model establishes correlations between local lesions and diagnostic narratives. \textbf{(4) MLLM Clinical Reasoning (Fig.~\ref{fig1}(d)):} Integrated with LLMs, our visual encoder mitigates anatomic misidentification (e.g., kidney vs. liver structures) and enables medical vision-language QA.

In summary, MedP-CLIP improves CLIP's fine-grained spatial understanding through a feature-level region prompt integration mechanism. It seamlessly processes various prompts while preserving critical global context. Pre-trained on a meticulously curated large-scale dataset comprising 97.3 million medical region-text pairs, it achieves robust cross-modal fine-grained semantic reasoning capabilities. MedP-CLIP outperforms benchmark models across various medical tasks, including but not limited to zero-shot recognition, interactive segmentation, and empowering multimodal LLMs, thereby establishing an expressive and scalable foundation for biomedicine.

\section{Related Work}
\subsection{Medical VLMs}
CLIP~\citep{CLIP} has established powerful transferable representations through large-scale image-text alignment, but its application in the medical domain remains challenging due to the lack of prior anatomical knowledge and limitations in fine-grained semantic representation. To address this, domain-specific adaptation methods have been pre-trained on medical corpora. For instance, MedCLIP~\citep{wang2022medclip} and PMC-CLIP~\citep{lin2023pmc} leverage medical image-text data for specialized pre-training, enabling CLIP to align global information in medical images with relevant text. Further, BiomedCLIP~\citep{zhang2025multimodal} underwent large-scale training on the PMC-15M dataset containing 15 million biomedical image-text pairs, even outperforming state-of-the-art radiology-specific models on tasks like RSNA pneumonia detection. 
Recently, Khattak et al. developed UniMed-CLIP~\citep{khattak2024unimed} on datasets spanning six medical modalities, enhancing CLIP's multimodal generalization ability in medicine. More recent efforts have further expanded this direction: BIOMEDICA~\citep{lozano2025biomedica} constructs a large-scale open biomedical image-caption archive from scientific literature, MedicalNarratives~\citep{ikezogwo2025medicalnarratives} introduces localized medical narratives by aligning medical visual content with explanatory language and spatial traces, and SurgLaVi~\citep{perez2026surglavi} builds a large-scale hierarchical surgical vision-language dataset with semantically enriched clip-caption pairs for surgical representation learning. Meanwhile, CLIP-Guided Generative Network~\citep{zhang2025clip} explores CLIP-guided pathology nuclei image augmentation to improve nuclei segmentation and classification under limited annotations. Nevertheless, existing medical VLMs still largely lack a native and flexible region-prompt interface for clinician-in-the-loop interaction, which may be insufficient for diagnostic workflows requiring precise localization of anatomical structures and pathological regions.
Consequently, they are unable to support clinician-in-the-loop interaction requiring dynamic region-aware prompting, highlighting an unresolved gap in medical VLMs.

\subsection{Region-Aware Methods in VLMs}
Research on region-aware visual language models aims to address the limitations of global image representations. Early input-level adaptive methods, such as ReCLIP~\citep{ReCLIP} and CircleCLIP~\citep{circleCLIP}, guided model attention by cropping or overlaying regions of interest. However, such disruptive operations inevitably sacrifice critical contextual information, which is particularly detrimental in medical imaging where pathological structures exhibit complex spatial dependencies. Subsequent feature-level methods, including MaskCLIP~\citep{MaskCLIP} and MaskAdaptedCLIP~\citep{MaskAdaptedCLIP}, preserved structures through attention masking but severely lost edge information relevant to diagnosis. While RegionCLIP~\citep{zhong2022regionclip} improved box-level alignment and Alpha-CLIP~\citep{sun2024alpha} introduced auxiliary alpha channels for soft modulation, they respectively suffered from coarse spatial reasoning and an architectural reliance on clinically impractical high-precision mask annotations during inference. Recent biomedical CLIP extensions have also improved concept-level interpretability and cross-domain generalization, as CDCLIP~\citep{li2026cdclip} decomposes diagnostic categories into human-interpretable concepts for zero-shot medical image classification, while MMKD-CLIP~\citep{wang2025unifying} distills knowledge from multiple biomedical CLIP teachers to build a more generalist foundation model across modalities. In medical imaging, uncertainty-aware diagnostic phrase identification and grounding ~\citep{zou2025uncertainty} has further explored the alignment between clinically meaningful diagnostic expressions and corresponding image regions, emphasizing the importance of reliable region-level medical image-text correspondence for diagnostic interpretation. Expansion in the medical domain has demonstrated the utility of spatial priors: R-LLaVA~\citep{chen2025r} enhanced region awareness by directly injecting bounding boxes into image space for visual question answering, while MedPLIB~\citep{huang2025towards} supported flexible region prompts through multi-stage mixture-of-experts training. However, these methods either extended the global CLIP backbone in a task-specific manner (e.g., R-LLaVA) or required complex multi-stage training and inference pipelines (e.g., MedPLIB), thereby imposing significant limitations on scalability and practical deployment. In contrast, our MedP-CLIP addresses this issue by treating region prompts as a flexible region-prompt interface for interaction. It supports multiple spatial input modes without relying on external modules or post-hoc inference mechanisms, enabling precise and direct region-level representation within a unified architecture.

\subsection{Non-Native Region Integration in Medical VLMs}

The imperative for spatial precision in medical diagnosis has spurred the development of models capable of directly interpreting and responding to region-specific prompts. Such models aim to fuse textual instructions with localized image semantics to simulate clinician interaction paradigms. For instance, MedRegA~\citep{medregaref} leverages sample-specific lesion or anatomical structure coordinates to focus attention, while ~\citep{gaoref} integrate abnormal region awareness within multimodal fusion for generating more detailed descriptions. Although these methods surpass global models in localization capability, they frequently depend on multi-stage alignment processes, constraining model flexibility. Concurrently, techniques like M2CRL~\citep{m2crlref} employ intricate masking schemes for endoscopic video analysis, and VertFound~\citep{vertfoundref} combines CLIP semantics with spatial features from SAM~\citep{sam} for fine-grained vertebra classification; however, these approaches are often confined to specific tasks or modalities. Beyond region-aware vision-language modeling, recent studies have explored uncertainty-aware segmentation~\citep{zou2025uncertainty} and uncertainty-guided multi-prompt adaptation~\citep{zhou2025medsam} to improve the reliability of prompt-based medical segmentation systems, further highlighting the importance of robust prompt-region interaction in medical image analysis. Furthermore, MedDAM~\citep{meddamref} adapts local description models to prioritize clinically critical lesion features. In parallel, FedMedCLIP~\citep{wu2026federated} adapts CLIP to privacy-preserving heterogeneous medical classification by freezing CLIP encoders and communicating lightweight masked feature adapters, but its objective remains image-level classification rather than native region-prompt interaction. Critically, a common limitation persists: these solutions typically incur increased model complexity and computational overhead. More fundamentally, they rely on non-intuitive prompts (e.g., predefined coordinate systems) or auxiliary modules for spatial encoding. This dependence on specialized, non-native interfaces can limit seamless, clinician-centric interactions driven by intuitive prompts. Therefore, addressing this gap necessitates a model that natively supports intuitive region prompting. Our proposed MedP-CLIP is designed to accurately interpret regions of interest specified via intuitive clinician prompts or perception models, establishing native region-level perception within the CLIP for medical scenarios at minimal modification cost.

\section{Method}
MedP-CLIP enhances the CLIP framework into a region-promptable VLM. While preserving CLIP's inherent strength in global image-text alignment, MedP-CLIP uniquely enables region-specific visual reasoning conditioned on fine-grained spatial prompts, addressing the critical need for Medical RoIs analysis. To achieve this, we introduce two key components: (1) Large-scale Medical Region-Text Dataset: We compile the largest medical region-text dataset containing 97.3M pairs of localized image regions and corresponding textual descriptions. (2) Plug-in Prompt Modulation: We devise a lightweight, plug-in prompt modulation mechanism that seamlessly integrates clinician-specified spatial prompts (points, bounding boxes, or masks) into CLIP's feature embedding space, dynamically guiding the model's attention to target RoI regions. The CLIP vision encoder is trained end-to-end with the prompt integration module, while the proposed design preserves the original ViT backbone structure and introduces a lightweight region prompt integration mechanism at the feature level, thereby retaining broad compatibility with downstream medical applications that benefit from CLIP’s transferability.

\subsection{Medical Image Region–Text Pairs Generation}\label{sec:3.1}
\begin{figure*}[pos=t] 
	\centering 
	\includegraphics[width=\textwidth]{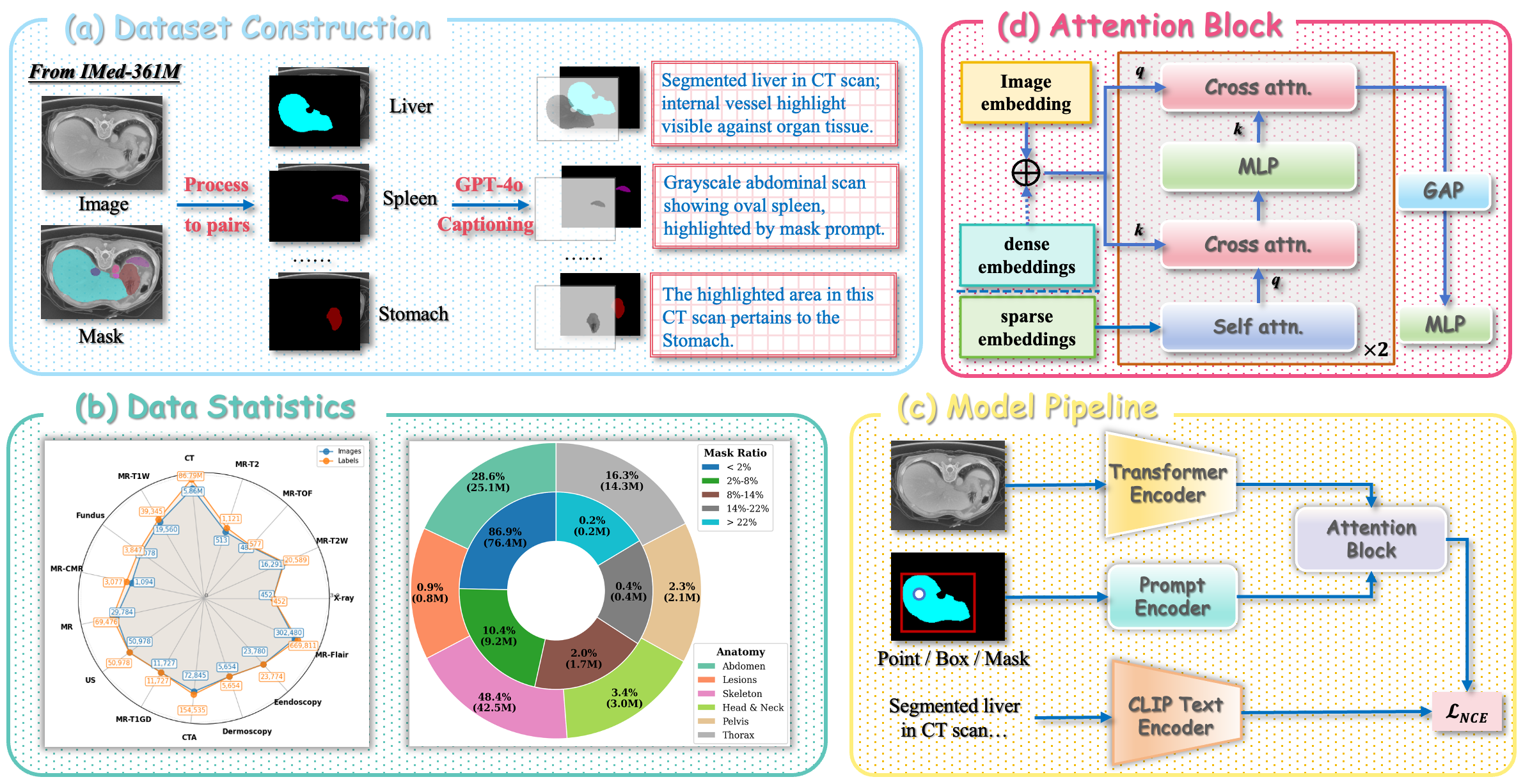} 
	\caption{MedP-CLIP: data, architecture and training. From IMed-361M we convert pixel-level organ/lesion masks into region–text pairs by generating prompts (point/box/mask) and GPT-4o based clinical descriptions (a). Dataset statistics across modalities, anatomies and mask ratios (b). Model pipeline: the image is encoded by a ViT, while the prompt is embedded by a lightweight Prompt Encoder (c). A feature-level Attention Block fuses dense and sparse prompt embeddings with image tokens via cross/self attention (d).}
	\label{fig2}
\end{figure*}

As shown in Fig.~\ref{fig2}(a), we generate medical image region-text pairs based on the IMed-361M dataset~\citep{cheng2025interactive}. This dataset was originally designed for interactive segmentation, providing pixel-level mask annotations and category labels for 6.4M medical images. To construct region-text pairs, we first decompose each original image into $n$ region-annotated samples, where $n$ corresponds to the number of different anatomical structures in the image.

The core of our data generation pipeline leverages GPT-4o to synthesize clinically based textual descriptions. For each triplet $\{I, M_k, C_k\}$ (representing the original image, the k-th region mask, and the associated category label), we input this information into GPT-4o. The model is given constrained instructions to generate comprehensive descriptions covering both the global context and specific features of the masked region, as well as to produce different variants based on provided templates. Example descriptions include: ``The contrast-enhanced CT image shows the segmented liver parenchyma, with prominent vascular enhancement within the masked region'' and ``The grayscale abdominal scan displays the oval-shaped splenic parenchyma highlighted by the overlaid mask''. Comprehensive implementation details of the text generation framework and our quality control mechanisms are elaborated in App.~ \ref{sup_qcm}.

As shown in Fig.~\ref{fig2}(b), we ultimately obtained a dataset of 97.3M region-text pairs, covering 15 different modalities, significantly surpassing the 6 modalities used in UniMed-CLIP pre-training. The region-text pairs were systematically categorized into six major anatomical groups (abdomen, bone, head neck, pelvis, thorax, and lesions), comprehensively covering the primary human anatomical systems. Crucially, quantitative analysis revealed that 86.9\% of the regions accounted for less than 2\% of the total image area, empirically demonstrating the prevalence of highly localized RoIs in medical imaging.

\subsection{MedP-CLIP}\label{sec:3.2}

MedP-CLIP bridges the gap between user-provided spatial prompts and medical image understanding through a unified framework that processes visual data ($I$), regional text descriptions ($T$), and prompts ($P$). As depicted in Fig.~\ref{fig2}(c), our architecture enhances CLIP's foundational capabilities with explicit region-aware reasoning via an attention block that jointly reasons over visual features and prompt semantics. This enables precise alignment between clinician-specified regions and textual descriptions.

\subsubsection{Image and Text Encoder}
We adopt CLIP's Vision Transformer (ViT) architecture as the image encoder, specifically instantiated as ViT-B/16@224 and ViT-L/14@336. Unless otherwise specified, the CLIP vision encoder is trained end-to-end together with the prompt integration module, while the CLIP text encoder is kept frozen. The vision encoder processes the input image $I$ into a sequence of patch tokens $\mathbf{F_I}\in\mathbb{R}^{L\times D_1}$, where $L$ denotes the number of patches and $D_1$ represents the hidden feature dimension. Meanwhile, the CLIP text encoder encodes the regional text description $T$, producing a global text feature $\mathbf{F}_T \in \mathbb{R}^{D}$, where $D$ represents the feature dimension.

\subsubsection{Prompt Encoder} Inspired by SAM~\citep{sam}, our prompt encoder processes two types of prompt information: sparse prompts (points, boxes) and dense prompts (masks). For sparse prompts, we combine positional encoding and learnable type embeddings to represent each point or box, forming sparse prompt features $\mathbf{F_{Sparse}}\in\mathbb{R}^{S\times D_1}$, where S denotes the number of sparse prompts. For dense prompts, we employ a set of lightweight convolutional networks for feature embedding, generating dense prompt features $\mathbf{F_{Dense}}\in\mathbb{R}^{L\times D_1}$ that maintain spatial resolution consistency with the image patch tokens.

\subsubsection{Attention Block} 
As shown in Fig.~\ref{fig2}(d), this block is the core component of MedP-CLIP, leveraging the geometric positions and type information of prompts to guide the model in focusing on user-specified image regions. It primarily consists of the following sequential processing steps: (1) Self-attention computation for sparse embeddings, and element-wise addition of dense embeddings (if present) with image embeddings; (2) Use sparse embeddings as queries to perform cross-attention computation with image embeddings, followed by a pointwise MLP update for each embedding; (3) Use image embeddings as queries to perform cross-attention computation with sparse embeddings; (4) Obtain the final image embedding $\mathbf{F_A}\in\mathbb{R}^{D}$ through $GAP$ and $MLP$. During computation, each self-attention/cross-attention module and $MLP$ includes residual connections and layer normalization layers. The above process can be expressed as follows:

\begin{align}
\left\{
\begin{aligned}
&\mathbf{F_A}' = {SelfAttn}(\mathbf{F_{Sparse}}), \mathbf{F_I}' = \mathbf{F_{Dense}} + \mathbf{F_{I}} \\
&\mathbf{F_A}'' = {MLP}({CrossAttn}(\mathbf{F_A}', \mathbf{F_I}')) \\
&\mathbf{F_A}''' = {CrossAttn}(\mathbf{F_I}', \mathbf{F_A}'') \\
&\mathbf{F_A} = {MLP}({GAP}(\mathbf{F_A}'''))
\end{aligned}
\right.
\end{align}

\subsubsection{Training Objective}\label{sec:training}
Given a batch of $N$ triplets $\{I_i,P_i,T_i\}^{N}_{i=1}$ where each triplet consists of an image, a user prompt, and a region-specific text description, we first normalize the output features from the attention module $\mathbf{F_A}$ and the text encoder $\mathbf{F_T}$ to obtain normalized embeddings $\hat{\mathbf{F_A}}$ and $\hat{\mathbf{F_T}}$. We then compute the following objectives including similarity matrix calculation, bidirectional contrastive losses, and the final symmetric loss function:

\begin{align}
\begin{cases}
 \mathcal{{L}_{I\rightarrow T}}=\frac{-1}{N}\sum_{i=1}^{N} \log  
\frac{\exp\big(\langle \hat{\mathbf{F}}_{A,i}, \hat{\mathbf{F}}_{T,i} \rangle / \tau\big)}
{{\textstyle \sum_{k=1}^{N}}  \exp\big(\langle \hat{\mathbf{F}}_{A,i}, \hat{\mathbf{F}}_{T,k} \rangle / \tau\big)} \\
\mathcal{{L}_{T\rightarrow I}}=\frac{-1}{N}\sum_{j=1}^{N}\log
\frac{\exp\big(\langle \hat{\mathbf{F}}_{T,j}, \hat{\mathbf{F}}_{A,j} \rangle / \tau\big)}
{{\textstyle \sum_{k=1}^{N}} \exp\big(\langle \hat{\mathbf{F}}_{T,k}, \hat{\mathbf{F}}_{A,j} \rangle / \tau\big)} \\
\mathcal{{L}}_{{NCE}}=\frac{1}{2}(\mathcal{L}_{I\rightarrow T} + \mathcal{L}_{T\rightarrow I})
\end{cases}
\end{align}

where $\tau$ is a learnable temperature. It should be noted that while our ground-truth data is derived from precise pixel-level masks, we actively simulate degraded, sparse inputs during training to decouple the model from high-precision mask dependency. Specifically, to prevent the model from over-relying on any single prompt type while maintaining holistic recognition capabilities, for each image we randomly select one prompt type from point prompts, box prompts, mask prompts, and point+box combined prompts with probabilities $\{0.3, 0.3, 0.3, 0.1\}$. This strategy forces the model to map imprecise spatial inputs to complex semantic regions, thereby ensuring robust native promptability during inference. Additionally, with a small probability ($p=0.1$), prompts are entirely discarded to train on complete image-text pairs.

\subsubsection{MedP-CLIP for Downstream Tasks}
Following pre-training, MedP-CLIP enables versatile downstream capabilities: (1) image-level zero-shot recognition with optional prompt customization; (2) region-level retrieval and classification within medical images using diverse prompt formats (points, bounding boxes, or combinations); (3) serving as the backbone for robust interactive medical image segmentation when integrated with the Segment Anything Model (SAM); and (4) plug-and-play replacement of existing visual encoders within medical large language model pipelines, requiring no architectural modifications.

\section{Experiments}
\subsection{Training Setup} 
All MedP-CLIP variants were trained on 16 NVIDIA RTX 4090 GPUs (24GB). The hyperparameters listed in Tab.~\ref{sup_hyper} were consistently applied across all pre-training experiments.

\begin{table}[ht]
\centering
\caption{Hyperparameter configurations for MedP-CLIP pre-training}
\label{sup_hyper}
\setlength{\tabcolsep}{28pt} 
\begin{tabular}{ll} 
\toprule[1.5pt]
\textbf{Hyperparameters} & \textbf{Value} \\
\midrule
{Optimizer} & AdamW~\citep{loshchilov2017decoupled} \\
{Learning rate} & {$1e-2$} \\
{Weight decay} & {$1e-2$} \\
{Optimizer momentum} & $\beta_1=0.9,\ \beta_2=0.999$  \\
{Learning rate schedule} & {MultiStepLR}, Step=\{25, 50, 75, 90, 120\} \\
{Epochs} & 150 \\
{Warmup (in steps)} & 800 \\
{Random seed} & 2025 \\
{Image mean} & (123.675, 116.28, 103.53) \\
{Image std} & (58.395, 57.12, 57.375) \\
{Augmentation} & RandScaleIntensityd, RandShiftIntensityd~\citep{cardoso2022monai} \\
{Log scale} & 4.6052 \\
{Backbones} & ViT-B/16@224, ViT-L/14@336~\citep{CLIP} \\
\bottomrule[1.5pt]
\end{tabular}

\end{table}

\subsection{Evaluation Dataset and Metrics}

To rigorously evaluate MedP-CLIP across the four downstream tasks, including zero-shot image-level classification, region-level retrieval, interactive segmentation, and medical visual question answering (VQA), we conduct experiments on a diverse and comprehensive benchmark comprising 11 publicly available medical imaging datasets spanning 8 modalities (CT, MRI, fundus, X-ray, etc.) and 129,298 expert-annotated samples. As shown in Tab.~\ref{datasets_overview}, all evaluation datasets are carefully held out from the pre-training corpus to ensure a strict assessment of out-of-distribution generalization under realistic clinical conditions. For AMOS2022~\citep{ji2022amos}, we explicitly separated the dataset at the patient level before region-text pair construction. Cases assigned to downstream evaluation were completely excluded from the GPT-4o-based text generation process and from the MedP-CLIP pre-training corpus. Therefore, no patient, image, mask, or region-text pair from the AMOS2022 evaluation subset overlaps with the pre-training data, preventing patient-level and image-level leakage.
Specifically, for zero-shot classification, we use COVID-CT~\citep{yang2020covid}, ACL~\citep{bien2018mrnet}, ChestCT~\citep{matsuyama2024performance}, and ACRIMA~\citep{ovreiu2021deep}. Region-level retrieval is evaluated on Br35H~\citep{wageh2024brain}, AMOS2022~\citep{ji2022amos}, and 3D-IRCADB~\citep{soler20103d}. Interactive segmentation is benchmarked on ISLES~\citep{hernandez2022isles}, SegThor~\citep{lambert2020segthor}, and TotalSegmentator MRI~\citep{d2024totalsegmentator}. For medical VQA, we adopt MeCoVQA-R~\citep{huang2025towards}, the most comprehensive region-aware benchmark to date, featuring 126,661 training and 2,637 test questions across eight modalities.

\begin{table}[pos=ht]
  \centering
  \caption{Overview of evaluation datasets for image-level classification, region-level retrieval, and interactive segmentation}
  \label{datasets_overview}
  \begin{tabular}{llllll}
    \toprule[1.5pt]
    \textbf{Dataset} & \textbf{Modality} & \textbf{Classes} & \textbf{Evaluation type}& \textbf{Image Num.} & \textbf{Mask Num.}\\
    \midrule
    Covid-CT~\citep{yang2020covid}        & CT     & 2  & Image-Level Class. & 746    & None       \\
    ACL~\citep{bien2018mrnet}             & MRI    & 2  & Image-Level Class. & 1021   & None        \\
    ChestCT~\citep{matsuyama2024performance}         & CT     & 4  & Image-Level Class. & 1000   & None        \\
    ACRIMA~\citep{ovreiu2021deep}          & Fundus & 2  & Image-Level Class. & 705    & None      \\
    Br35H~\citep{wageh2024brain}           & MRI    & 2  & Region-Level Class. & 3000  & 801       \\
    AMOS2022~\citep{ji2022amos}        & CT     & 15 & Region-Level Class. & 19310 & 77822  \\
    3D-IRCADB~\citep{soler20103d}       & CT     & 16 & Region-Level Class. & 2823  & 12250   \\
    ISLES~\citep{hernandez2022isles}           & CT     & 2  & Interactive Seg.    & 364   & 364   \\
    SegThor~\citep{lambert2020segthor}         & CT     & 4  & Interactive Seg.    & 516   & 1258   \\
    TotalSeg. MRI~\citep{d2024totalsegmentator}   & MRI    & 12  & Interactive Seg.    & 2147  & 8100   \\
    
   \bottomrule[1.5pt]
  \end{tabular}

\end{table}

Evaluation metrics are aligned with task semantics and community standards: Top-1/Top-5 accuracy and recall for classification and region retrieval; Dice coefficient for interactive segmentation; and precision and recall for region-level VQA, following the MedPLIB protocol~\citep{huang2025towards}. This unified evaluation framework enables a holistic and clinically meaningful assessment of MedP-CLIP’s multimodal reasoning, spatial grounding, and zero-shot adaptability across heterogeneous medical imaging scenarios.

\begin{table*}[pos=!htbp]
	\centering
	\caption{Comparison of zero-shot image-level classification accuracy across four medical datasets}
 
	\setlength{\tabcolsep}{10pt}

	\begin{tabular}{l|ccccc} 
		\toprule[1.5pt]
		\textbf{Models} & \textbf{COVID-CT} & \textbf{ACL} & \textbf{ChestCT} & \textbf{ACRIMA} & \textbf{Average} \\
		\midrule
		CLIP \citep{CLIP}         & 48.28\% & 44.27\% & 26.59\% & 57.61\% & 44.19\% \\
		MedCLIP \citep{wang2022medclip}      & 47.78\% & 40.55\% & 25.45\% & 45.66\% & 39.86\% \\
		BiomedCLIP \citep{zhang2025multimodal}   & 62.07\% & 48.38\% & 22.19\% & 56.05\% & 47.17\% \\
		UniMed-CLIP \citep{khattak2024unimed}  & 61.08\% & 43.78\% & 35.40\% & \textbf{59.32\%} & 49.90\% \\
        ASG \citep{li2024anatomical}          & 53.18\%  & 55.73\%  & 32.14\% & {56.19\%} & {49.31\%} \\
        {CARZero \citep{lai2024carzero}}       & {59.06\%} & {49.36\%} & {36.70\%} & {51.07\%} & {49.05\%} \\
        {GenMedCLIP \citep{ikezogwo2025medicalnarratives}} & {60.24\%} & {55.83\%} & {\textbf{37.85\%}} & {56.61\%} & {52.63\%} \\
        {BMC-CLIP \citep{lozano2025biomedica}} & {65.88\%} & {54.16\%} & {26.75\%} & {56.33\%} & {50.78\%} \\
		\midrule
		\textbf{MedP-CLIP (ViT-B/16)}  & \textbf{68.47\%} & \textbf{59.35\%} & 33.12\% & 56.47\% & \textbf{54.35\%} \\
		\bottomrule[1.5pt]
	\end{tabular}
	\label{tab1}
\end{table*}

\begin{figure}[pos=t] 
	\centering 
	\includegraphics[width=0.85\textwidth]{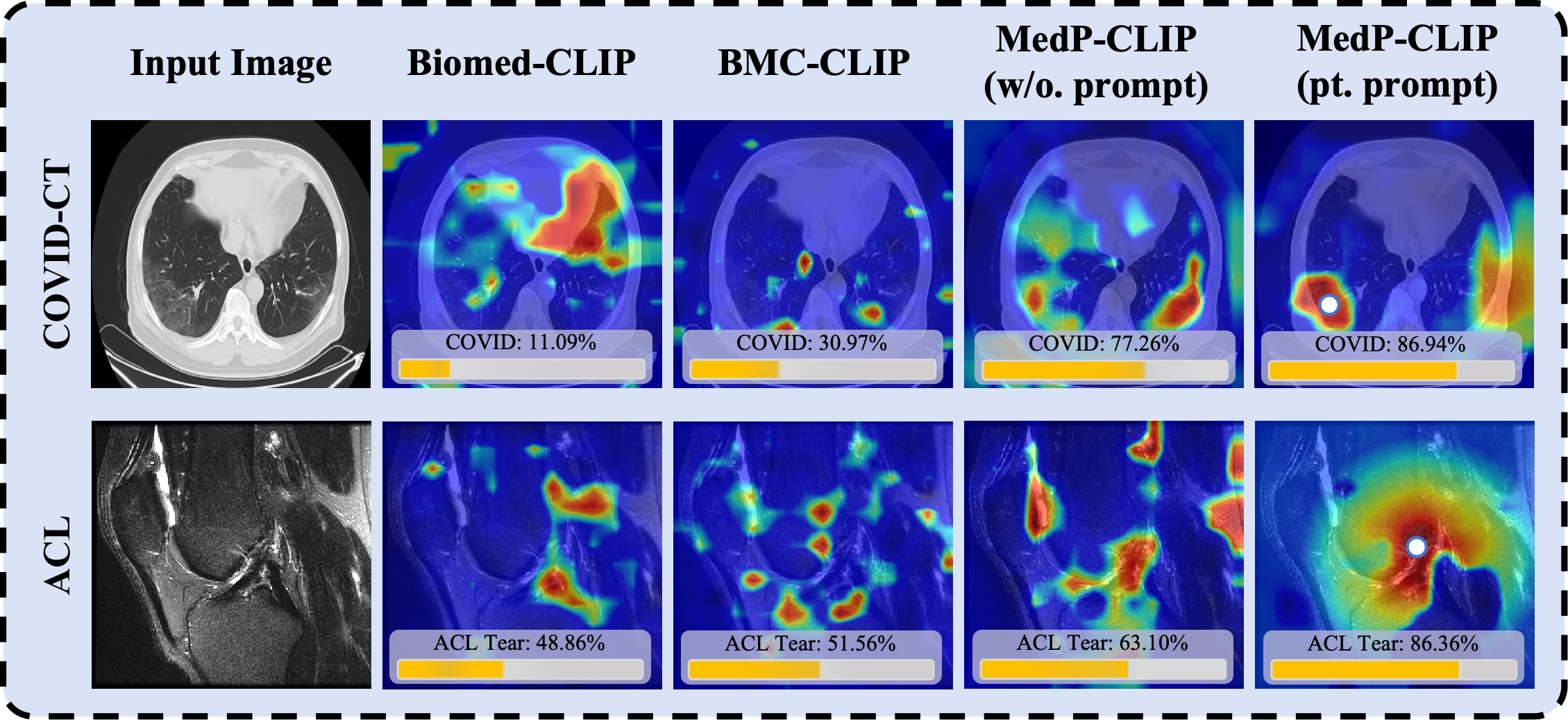} 
	\caption{{Qualitative comparison of attention maps for zero-shot classification on COVID-CT and ACL datasets. }}
	\label{fig3}
\end{figure}

\subsection{Image-Level Zero-Shot Classification}

As mentioned above, we conducted zero-shot classification experiments on four different medical imaging datasets: COVID-CT~\citep{yang2020covid}, ACL~\citep{bien2018mrnet}, ChestCT~\citep{matsuyama2024performance}, and ACRIMA\\ \citep{ovreiu2021deep}. Among them, ChestCT contains four categories (normal, adenocarcinoma, large cell carcinoma, and squamous cell carcinoma), while the other datasets are binary classification tasks. Although MedP-CLIP is primarily designed for region-level reasoning, our results demonstrate that it effectively retains the global representation capabilities inherent to CLIP and its recent medical variants. As shown in Tab.~\ref{tab1}, MedP-CLIP achieves the best performance on the COVID-CT and ACL datasets, with improvements of up to 3.52\% compared to the best existing methods. The model also maintains strong performance on the more challenging ChestCT dataset, which includes multiple disease categories that may benefit from our region-aware design. {Notably, MedP-CLIP achieves an average accuracy of 54.35\% across all datasets, representing a 1.72\% improvement over the current top-performing GenMedCLIP baseline.}

{To further corroborate these quantitative findings, we provide qualitative visual comparisons in Fig. \ref{fig3}. As illustrated, state-of-the-art global representation models, such as Biomed-CLIP and BMC-CLIP, often exhibit diffuse attention maps and struggle to localize spatially confined pathological regions. While MedP-CLIP without explicit prompts already demonstrates improved focal attention (e.g., yielding a 77.26\% confidence score on the COVID-CT sample compared to Biomed-CLIP's 11.09\%), the introduction of point prompts enables highly precise region-text alignment. By explicitly guiding the model to the target lesion, MedP-CLIP successfully mitigates irrelevant background interference, resulting in the most accurate attention maps and the highest diagnostic confidence scores. These results confirm that our method, while explicitly learning to handle regional prompts, does not compromise overall semantic alignment, making MedP-CLIP more versatile and reliable in practical clinical applications.}

\begin{table*}[pos=!htbp]
    \centering
    \footnotesize
    \caption{Region-level classification performance on the Br35H dataset under different prompt settings. Accuracy values of MedP-CLIP with prompts are reported as mean $\pm$ std over five runs, with the corresponding 95\% confidence intervals shown in brackets. Params. and FPS denote the number of model parameters and inference speed, respectively}
    
    \begin{tabular}{l|c|c|c|c|c|c|c} 
        \toprule[1.5pt]
        \textbf{Models} &
        \makecell{\textbf{Data Scale}} &
        \makecell{\textbf{Params.}} &
        \makecell{\textbf{FPS}} &
        \makecell{\textbf{Medical} \\ \textbf{Domain}} &
        \makecell{\textbf{Dataset} \\ \textbf{Reg. Info.}} & 
        \makecell{\textbf{Reg.-aware} \\ \textbf{Design}} & 
        \textbf{Br35H} \\
        \midrule
        
        CLIP \citep{CLIP}         
        & 400 M  
        & 150 M 
        & 275 
        & \xmark 
        & \xmark 
        & \xmark 
        & 52.93\% \\
        
        MedCLIP \citep{wang2022medclip}  
        & 0.6 M  
        & 136 M 
        & 277
        & \cmark 
        & \xmark 
        & \xmark 
        & 48.56\% \\
        
        BiomedCLIP \citep{zhang2025multimodal}   
        & 15 M  
        & 196 M 
        & 268 
        & \cmark 
        & \xmark 
        & \xmark 
        & 91.95\% \\
        
        UniMed-CLIP \citep{khattak2024unimed}  
        & 5.3 M 
        & 196 M 
        & 269 
        & \cmark 
        & \xmark 
        & \xmark 
        & 74.59\% \\
        
        ASG \citep{li2024anatomical}          
        & 217 K  
        & 291 M  
        & 566 
        & \cmark 
        & \cmark 
        & \cmark 
        & 67.03\% \\
        
        CARZero \citep{lai2024carzero}   
        & 377 K  
        & 221 M 
        & 148 
        & \cmark 
        & \xmark 
        & \cmark 
        & 51.93\% \\
        
        GenMedCLIP \citep{ikezogwo2025medicalnarratives}   
        & 4.7 M  
        & 150 M 
        & 280 
        & \cmark 
        & \cmark 
        & \xmark 
        & 78.93\% \\
        
        BMC-CLIP \citep{lozano2025biomedica}     
        & 24 M   
        & 428 M 
        & 207  
        & \cmark 
        & \xmark 
        & \xmark 
        & 91.70\% \\
        
        \midrule
        
        \makecell[l]{MedP-CLIP (ViT-B/16, w/o. prompt)}
        & \multirow{7}{*}{6.4 M}
        & \multirow{7}{*}{167 M}
        & 273
        & \multirow{7}{*}{\cmark}
        & \multirow{7}{*}{\cmark}
        & \multirow{7}{*}{\cmark}
        & 58.61\% \\
        
        \makecell[l]{MedP-CLIP (ViT-B/16, w. point)}
        & 
        & 
        & 269
        & 
        & 
        & 
        &\makecell{69.79\% $\pm$ 0.14\% \\ $[69.62, 69.96]$}
        \\
        
        \makecell[l]{MedP-CLIP (ViT-B/16, w. box)}
        & 
        & 
        & 269
        & 
        & 
        & 
        & \makecell{79.37\% $\pm$ 0.21\% \\ $[79.11, 79.63]$}\\

        \makecell[l]{MedP-CLIP (ViT-B/16, w. both)}
        & 
        & 
        & 269
        & 
        & 
        & 
        & \makecell{78.93\% $\pm$ 0.30\% \\ $[78.56, 79.30]$}\\
        
        \midrule
        
        \makecell[l]{\textbf{MedP-CLIP} \textbf{(ViT-L/14, w. both)}}
        & 6.4 M 
        & 455 M 
        & 71 
        & \cmark 
        & \cmark 
        & \cmark 
        & \makecell{\textbf{92.45\% $\pm$ 0.11\%} \\ $[92.31, 92.59]$}\\
        
        \bottomrule[1.5pt]
    \end{tabular}
    \label{tab:br35h}
\end{table*}

\subsection{Region-Level Retrieval and Classification}

In addition to global classification, we further evaluated region-level understanding on Br35H, a brain tumor classification benchmark that requires localized diagnostic reasoning. As shown in Tab.~\ref{tab:br35h}, MedP-CLIP without spatial prompts achieves 58.61\% accuracy, indicating that image-level representations alone are insufficient for accurately identifying localized pathological regions. Introducing lightweight spatial prompts substantially improves performance: the point prompt increases the accuracy to 69.79\% $\pm$ 0.14\%, while the bounding box prompt further raises it to 79.37\% $\pm$ 0.21\%, corresponding to a 20.76 percentage-point gain over the prompt-free setting. This result also surpasses the recent region-aware baseline GenMedCLIP by 0.44 percentage points. Interestingly, combining point and box prompts with the ViT-B/16 backbone yields 78.93\% $\pm$ 0.30\%, which is slightly lower than using the box prompt alone, suggesting that the bounding box already provides sufficiently precise regional constraints for this dataset and that additional point information may not always bring complementary benefits. The MedP-CLIP ViT-B/16 variants contain 167M parameters and run at 269 FPS when using prompts, which is comparable to CLIP, MedCLIP, BiomedCLIP, and UniMed-CLIP, while being substantially faster than CARZero and BMC-CLIP. This indicates that the performance gain is not obtained at the cost of heavy computational overhead. When scaling to the ViT-L/14 backbone, MedP-CLIP achieves the best accuracy of 92.45\% $\pm$ 0.11\%, outperforming strong medical vision-language baselines such as BiomedCLIP and BMC-CLIP by 0.50 and 0.75 percentage points, respectively. Moreover, the 95\% confidence interval of MedP-CLIP (ViT-L/14, w. both) is $[92.31, 92.59]$, whose lower bound remains higher than the accuracies of the strongest competing baselines, including BiomedCLIP and BMC-CLIP. This indicates that the observed improvement is stable across repeated runs rather than being caused by random variation. Although this larger variant introduces higher computational cost, with 455M parameters and 71 FPS, it provides a favorable high-accuracy configuration for scenarios where diagnostic performance is prioritized. The consistently small standard deviations across prompt-based variants further indicate stable performance over repeated runs.

\begin{figure}[pos=htbp] 
	\centering 
	\includegraphics[width=0.90\textwidth]{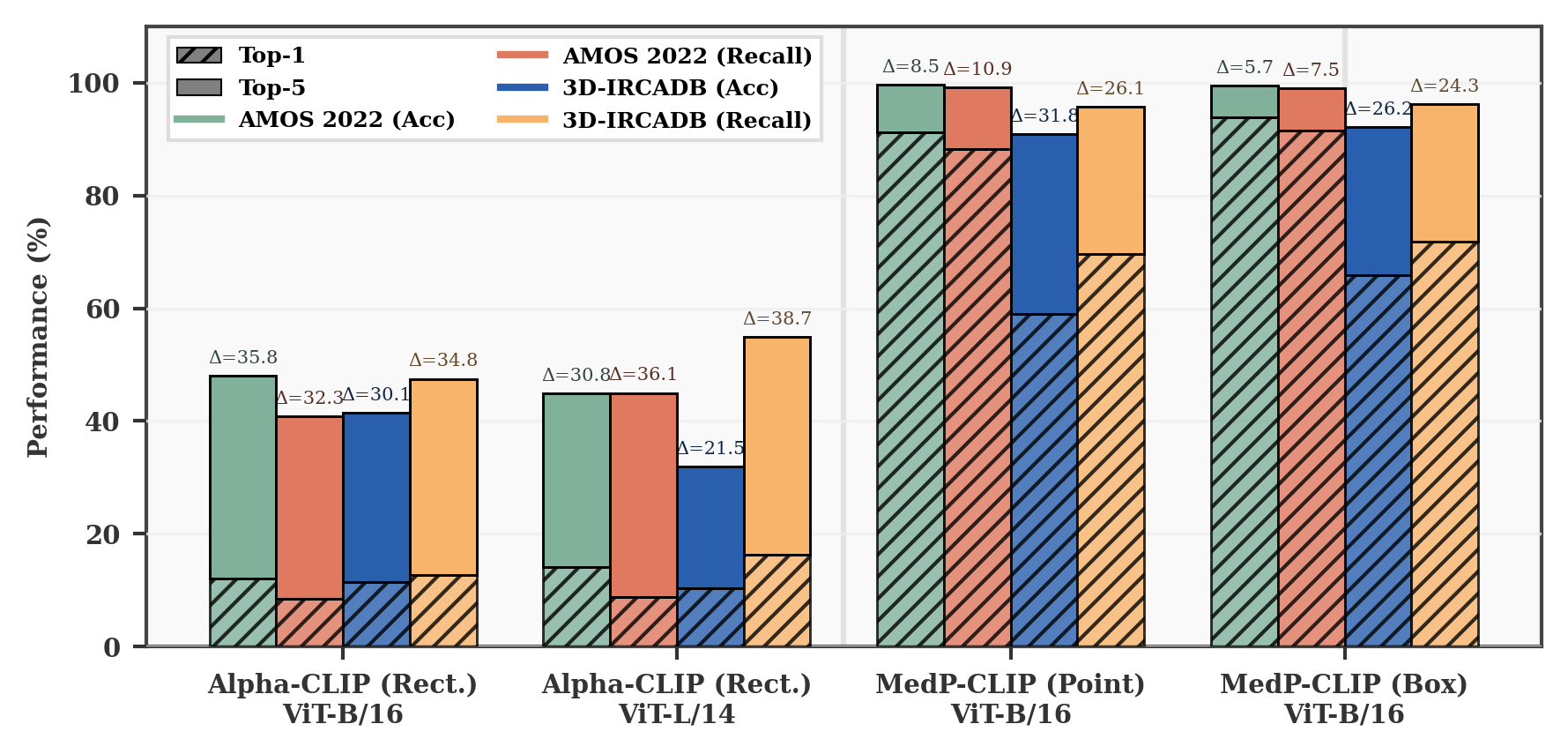} 
	\caption{Performance comparison of region-level classification on multi-organ datasets. ``$\Delta$'' denotes the performance difference between Top-5 and Top-1. ``Rect.'' indicates rectangular region prompts generated from original masks via Alpha-CLIP.}
	\label{fig4}
\end{figure}

We extended the evaluation to multi-organ matching tasks on AMOS2022 and 3D-IRCADB. Due to the lack of region-level tuning in medical baseline models, we benchmarked against Alpha-CLIP, a natural image foundation model. As shown in Fig.~\ref{fig4}, MedP-CLIP consistently outperformed Alpha-CLIP in region-aware tasks. Bounding box prompts, which provide explicit spatial context, generally yielded higher accuracy than point prompts, while composite prompts achieved the best overall accuracy and recall. To further examine the category-wise behavior, we provide a per-organ radar comparison on the 3D-IRCADB dataset.

\begin{figure}[pos=htbp] 
	\centering 
	\includegraphics[width=0.90\textwidth]{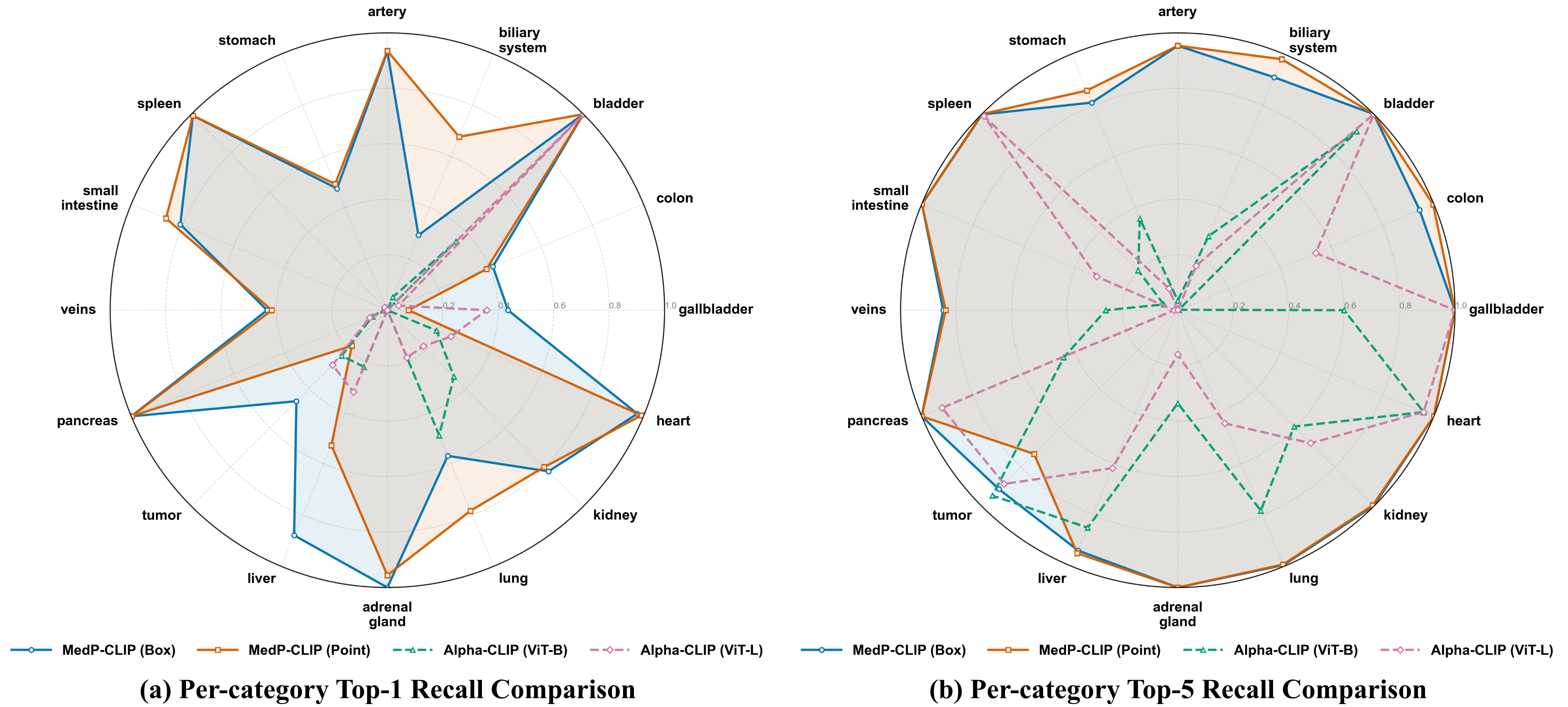} 
	\caption{Comparison of per-category Top-1 and Top-5 recall performance on the 3D-IRCADB dataset.}
	\label{supfig}
\end{figure}

\begin{figure}[pos=t] 
	\centering 
	\includegraphics[width=0.95\textwidth]{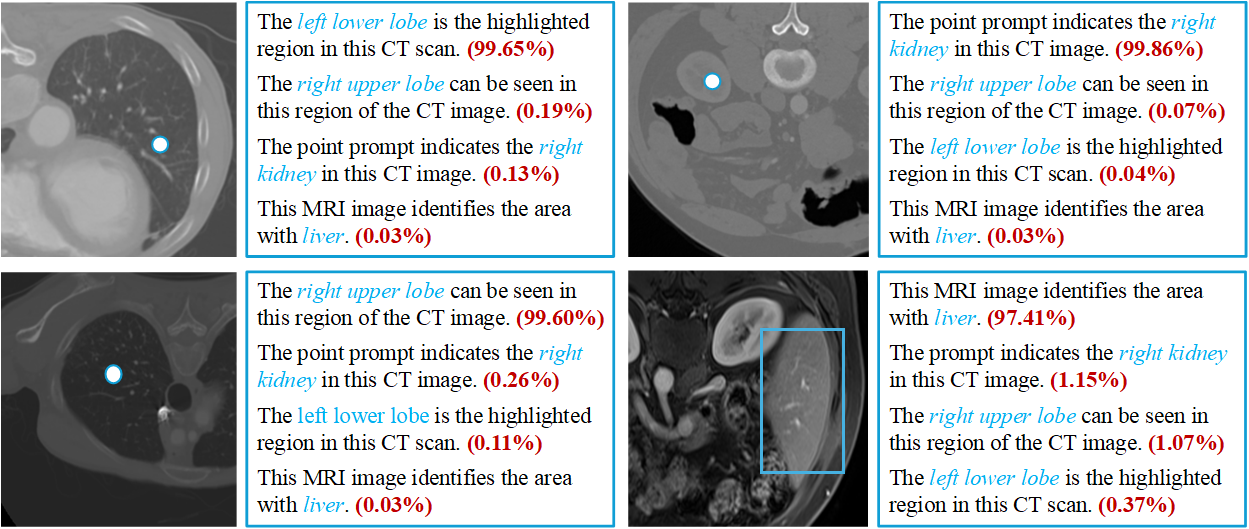} 
	\caption{Visual validation of MedP-CLIP's region-aware retrieval capabilities.}
	\label{fig5}
\end{figure}

As shown in Fig~\ref{supfig}, MedP-CLIP maintains consistently strong recall across most anatomical categories, including artery, bladder, heart, kidney, lung, adrenal gland, pancreas, spleen, and small intestine, where the performance is close to or reaches the outer boundary of the chart. In contrast, Alpha-CLIP exhibits highly uneven behavior across organs, with substantially degraded recall on several clinically important structures such as artery, biliary system, adrenal gland, veins, and small intestine. This clear per-class gap suggests that the advantage of MedP-CLIP does not merely come from improvements on a few dominant categories, but from more stable region-aware recognition across heterogeneous anatomical structures. Moreover, the comparison between point and box prompts reveals complementary behaviors: box prompts provide more reliable localization for spatially ambiguous or small structures, whereas point prompts can achieve comparable or slightly better performance for organs with clear visual boundaries. These fine-grained results further support the effectiveness of our region-aware prompt integration mechanism in handling diverse anatomical targets. Fig.~\ref{fig5} visually validates these findings: under flexible regional prompts, MedP-CLIP was able to retrieve semantically aligned text descriptions with near-certain confidence ($>$98\%). These findings confirm that: (i) our feature-level prompt fusion effectively integrates positional guidance; and (ii) prompt quality is directly correlated with discriminative localization capability, which is a key advantage for medical image interpretation.

\subsection{MedP-CLIP in Interactive Segmentation}

\begin{table}[!htbp]
\centering
\begin{minipage}[t]{0.52\linewidth}
    \centering
    \caption{Interactive segmentation results on external datasets}
    \setlength{\tabcolsep}{3pt} 
    \begin{tabular}{lccc} 
        \toprule[1.5pt]
        \multirow{2}{*}{\textbf{Models}} & \multicolumn{3}{c}{\textbf{Dataset / Dice (\%)}} \\ \cline{2-4}
        & ISLES & SegThor & Totalseg. \\
        \midrule
        SAM~\citep{sam} & 55.92 & 84.46 & 75.45\\
        SAM2~\citep{sam2} & 60.14 & 85.86 & 77.62 \\
        SAM-Med2D~\citep{cheng2023sam} & 68.22 & 86.43 & 75.92 \\
        IMIS-Net~\citep{cheng2025interactive} & 71.78 & \textbf{89.27} & 79.06 \\
        \midrule
        \textbf{MedP-CLIP+SAM} & \textbf{73.14} & 88.91 & \textbf{81.55} \\
        \bottomrule[1.5pt]
    \end{tabular}
    \label{tab4}
\end{minipage}
\hspace{0.02\linewidth} 
\begin{minipage}[t]{0.44\linewidth}
    \centering
    \caption{Region-level medical VQA on MeCoVQA-R}
    \begin{tabular}{lcc} 
        \toprule[1.5pt]
        \multirow{2}{*}{\textbf{Models}} & \multicolumn{2}{c}{\textbf{MeCoVQA-R (\%)}} \\ \cline{2-3}
        & Precision & Recall \\
        \midrule
        LLaVA \citep{LLaVA} & 18.01 & 44.98 \\
        LLaVA-Med \citep{li2023llava} & 15.47 & 34.86 \\
        LISA \citep{lai2024lisa} & 12.31 & 13.20 \\
        MedPLIB \citep{huang2025towards} & 64.92 & 63.84 \\
        \midrule
        \textbf{MedP-CLIP+LLaVA-Med} & \textbf{67.96} & \textbf{70.36} \\
        \bottomrule[1.5pt]
    \end{tabular}
    \label{tab5}
\end{minipage}
\end{table}

\begin{figure}[pos=htbp] 
	\centering 
	\includegraphics[width=0.95\textwidth]{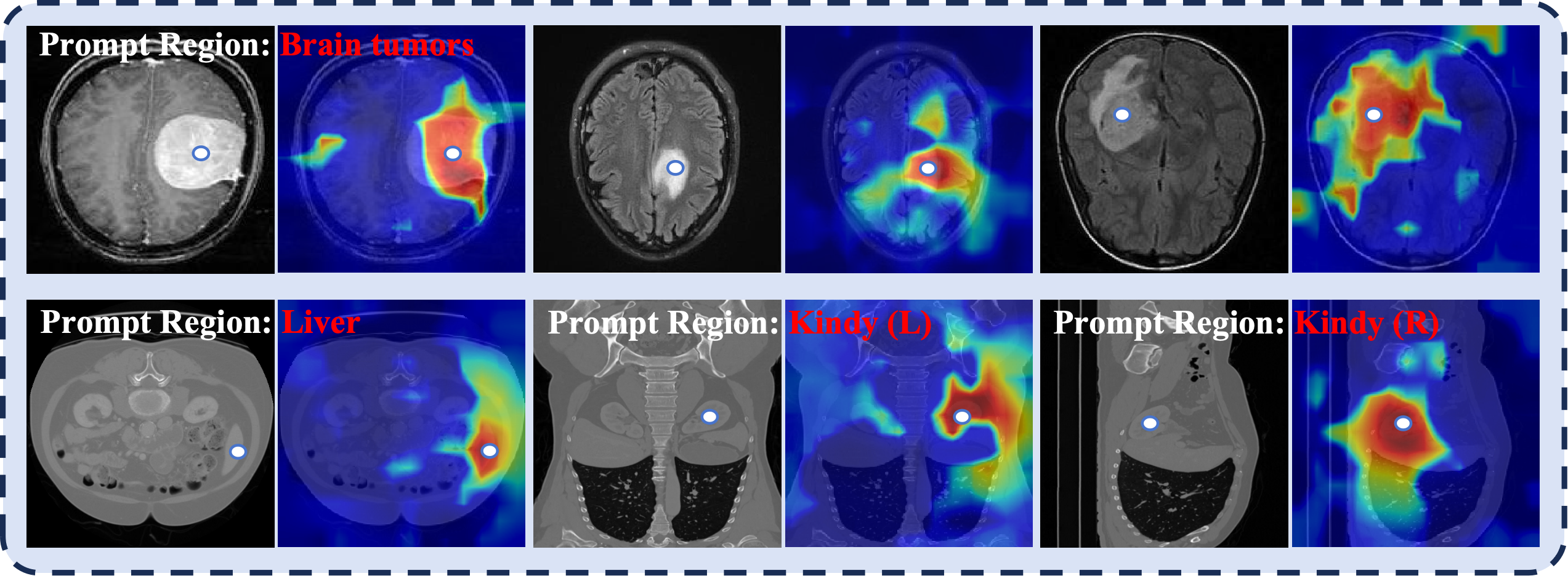} 
	\caption{Qualitative results of MedP-CLIP. Prompt guides the visual encoder to focus on the RoIs.}
	\label{fig6}
\end{figure}

Leveraging MedP-CLIP's intrinsic prompt affinity, we developed MedP-CLIP+SAM by substituting the original SAM ViT encoder. This architecture exploits synergy through shared prompt encoding features between MedP-CLIP (focusing the backbone on relevant regions) and SAM's mask decoder (guiding segmentation). Following the IMIS-Net~\citep{cheng2025interactive} protocol, we fine-tuned the decoder ($<$1\% of encoder parameters) for 5 epochs on IMed-361M, achieving rapid convergence. As detailed in Tab.~\ref{tab4}, MedP-CLIP+SAM delivers consistent and substantial improvements over the original SAM across all datasets. Critically, our model achieves new state-of-the-art performance on ISLES (73.14\%) and TotalSegmentator MRI (81.55\%), surpassing IMIS-Net. These results underscore MedP-CLIP's dual capability: it serves as a powerful backbone for downstream medical vision tasks while uniquely enabling targeted performance gains by directing computational resources towards clinically critical regions. Furthermore, visualization of the prompted feature responses (Fig.~\ref{fig6}) reveals that the model's attention consistently concentrates on prompt-relevant areas. This focused attention mechanism facilitates rapid localization of regions-of-interest for the segmentation model, contributing to the observed performance gains.

\begin{figure}[pos=!htbp]
  \centering
  \includegraphics[width=0.9\linewidth]{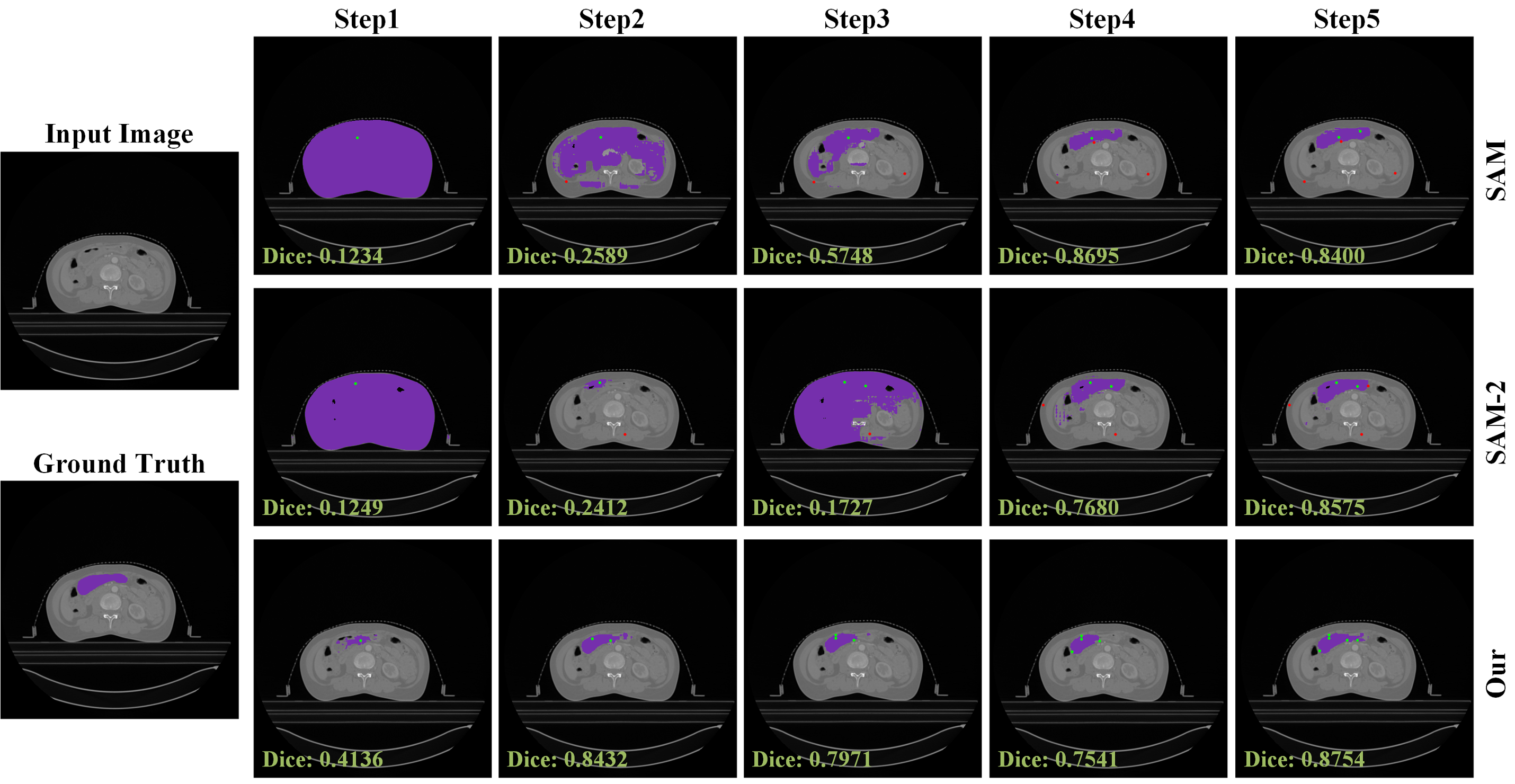}
  \caption{Segmentation results of MedP-CLIP+SAM with five points interaction.}
  \label{fig7}
\end{figure}

\begin{figure}[pos=t]
  \centering
  \includegraphics[width=0.9\linewidth]{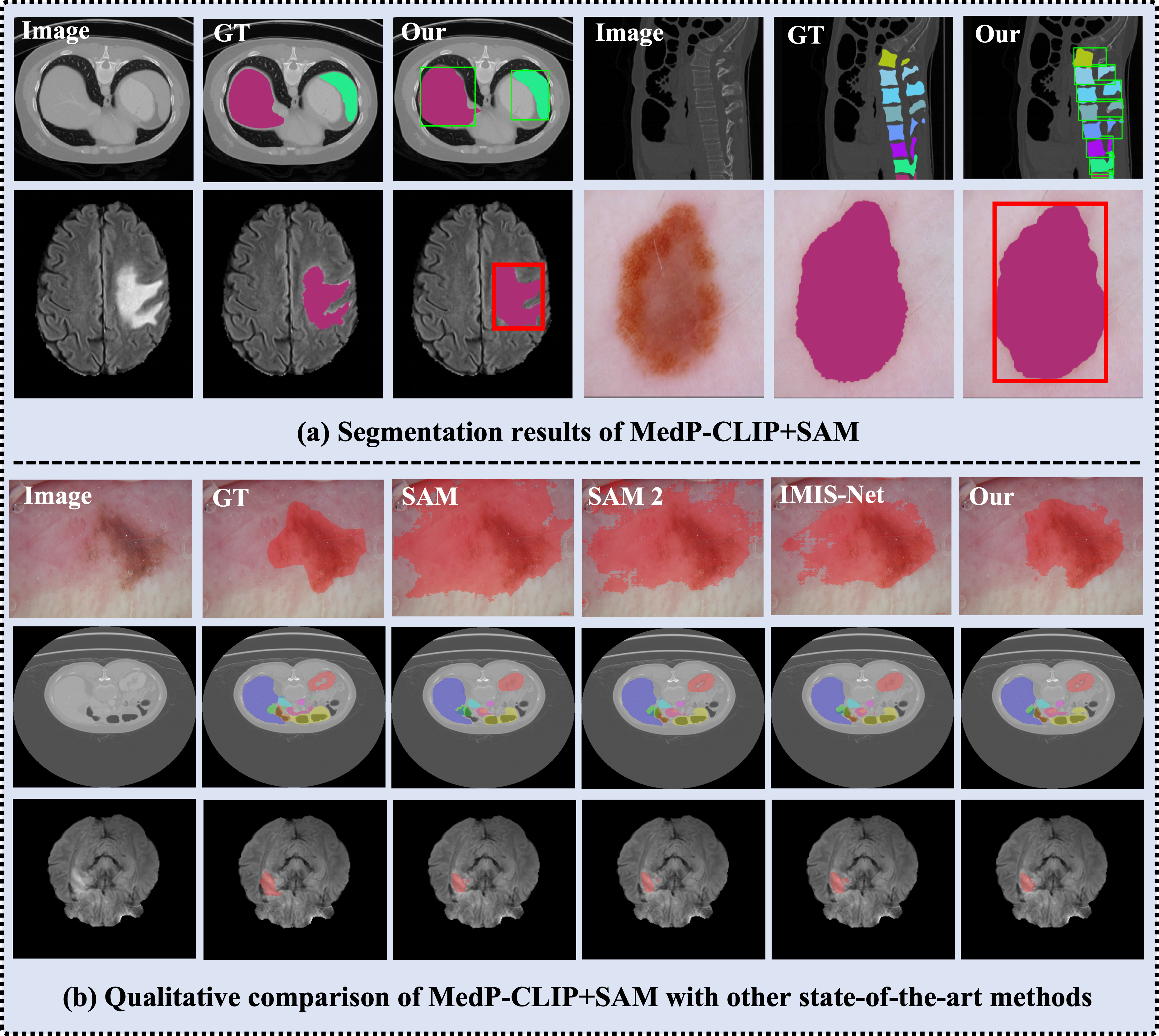}
  \caption{Qualitative experimental comparison of interactive segmentation.}
  \label{fig8}
\end{figure}

{In addition, we provide comprehensive qualitative comparisons in Fig.~\ref{fig7} and Fig.~\ref{fig8}. As illustrated in Fig.~\ref{fig8}(b), when benchmarked against established foundation models, baseline methods such as SAM and SAM 2 frequently struggle with ambiguous anatomical boundaries often exhibiting significant over-segmentation or failing to capture the complete region of interest (e.g., in complex skin lesions and multi-organ abdominal CTs). In contrast, MedP-CLIP+SAM, benefiting from its fine-grained region-aware capabilities, demonstrates a superior ability to accurately delineate complex structures and capture fine details across diverse modalities, generating masks that closely align with the ground truth. Additionally, as shown in Fig.~\ref{fig8}(a), our model seamlessly integrates with minimal user interactions, utilizing standard bounding boxes or a single click to effectively localize and segment various clinical targets.}

\subsection{Region-Level Medical VQA}

We adapt the LLaVA-Med~\citep{li2023llava} framework for region-level medical VQA by integrating our MedP-CLIP (ViT-B/16@224) as the vision encoder. This simple replacement enables focused reasoning on user-defined regions—specifically, bounding-box prompts derived from masks—while aligning projector dimensions via a two-layer MLP with GELU non-linearity. Following LLaVA-Med 1.5~\citep{li2023llava}, we retain Mistral-7B as the LLM, freeze the visual encoder, and fine-tune only the projector and LLM for one epoch on MeCoVQA-R to maintain efficiency. Evaluation on the MeCoVQA-R test split confirms that MedP-CLIP robustly directs attention to prompted regions, avoiding distractions from irrelevant image contexts seen in generic baselines (e.g., LLaVA and LISA). As summarized in Tab.~\ref{tab5}, MedP-CLIP sets a new state-of-the-art for region-level medical VQA, achieving 67.96\% precision and 70.36\% recall. This significantly outperforms recent approaches, including MedPLIB (64.92\%) and generic MLLMs such as LLaVA (18.01\%), LLaVA-Med (15.47\%), and LISA (12.31\%). The improvements stem from MedP-CLIP's enhanced encoding capabilities, which minimize error propagation in crowded medical scenes by focusing precisely on prompts. 

\begin{figure}[pos=!htbp]
  \centering
  \includegraphics[width=0.95\linewidth]{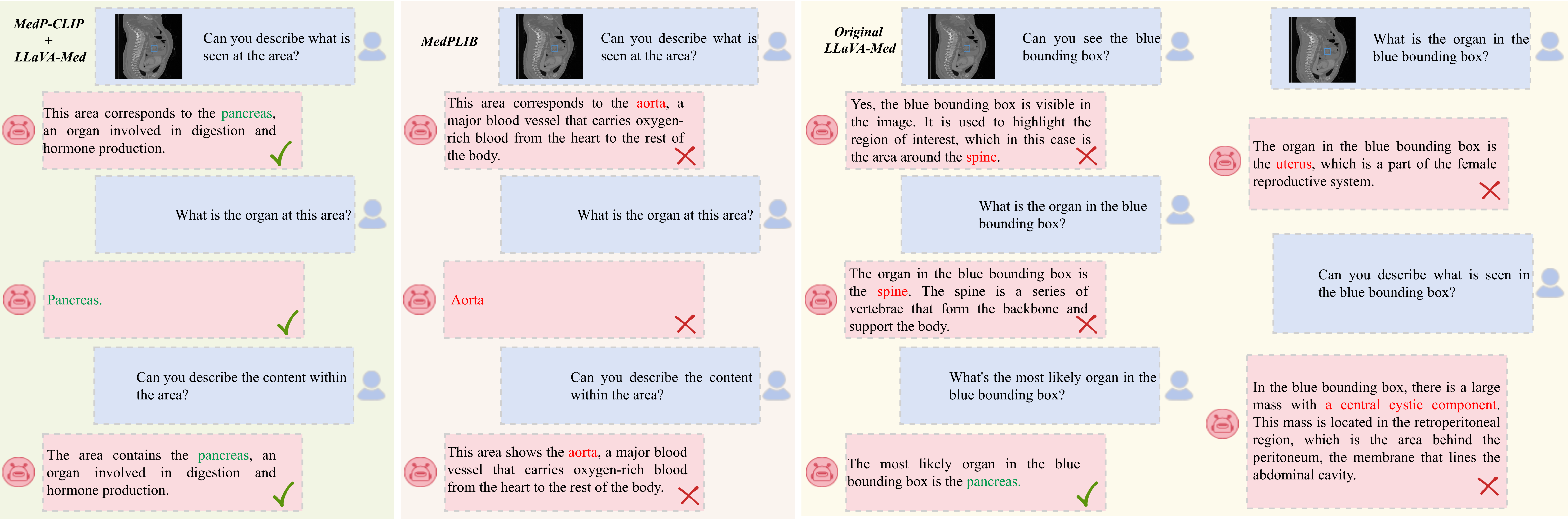}
  \caption{{Qualitative comparison of region-level medical Visual Question Answering (VQA) capabilities.}}
  \label{fig9}
\end{figure}

{Furthermore, to intuitively demonstrate the practical reasoning capabilities of our model, we provide qualitative comparisons in Fig. \ref{fig9}. When prompted with a specific bounding box highlighting the pancreas, baseline models such as MedPLIB and the original LLaVA-Med exhibit severe anatomical misidentification erroneously predicting the ``aorta'' or ``spine'' and generate hallucinated clinical findings. Conversely, MedP-CLIP+LLaVA-Med accurately grounds the visual prompt, correctly identifies the targeted organ, and provides clinically sound descriptions. These qualitative results explicitly corroborate our quantitative findings, proving that our feature-level region integration effectively mitigates error propagation in complex clinical scenarios.}

\subsection{{Case Study and Limitation Analysis}}

\begin{figure}[pos=!htbp]
  \centering
  \includegraphics[width=\linewidth]{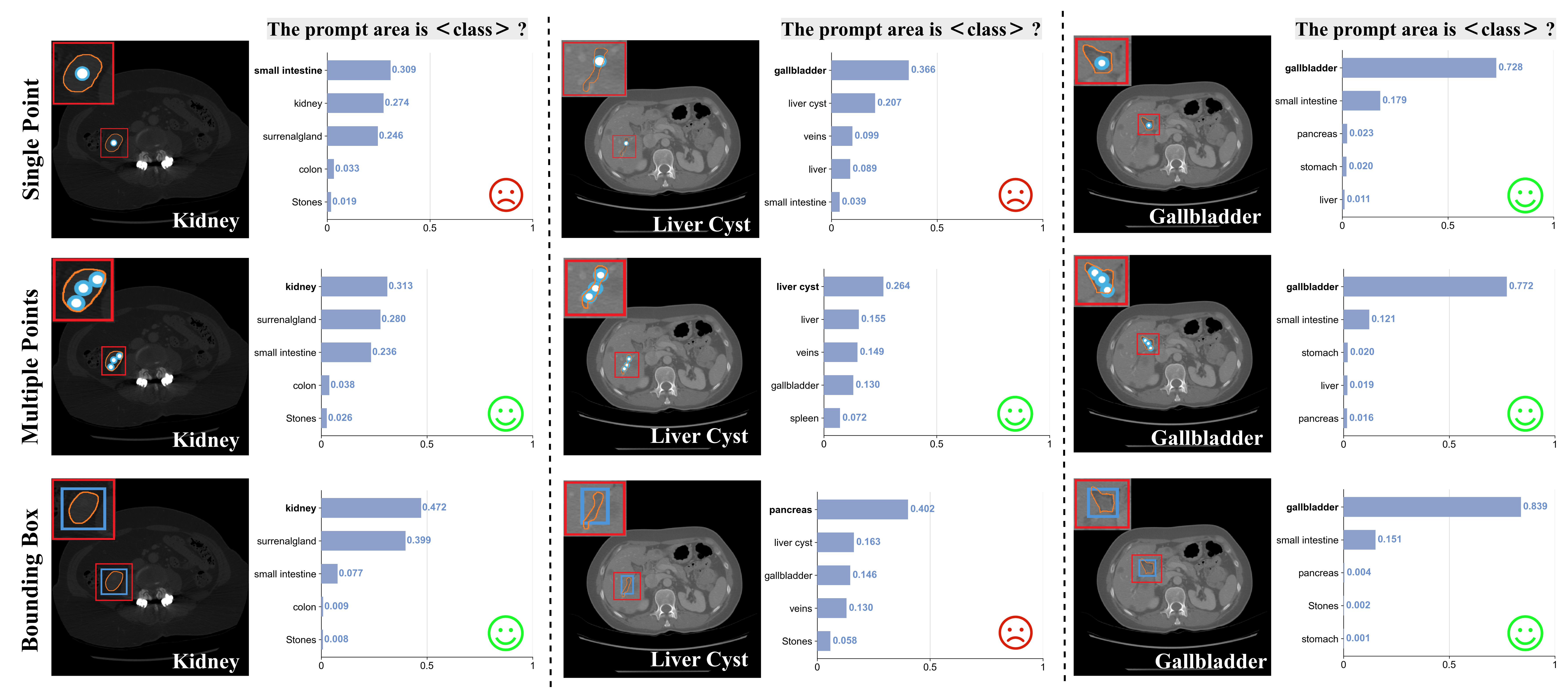}
  \caption{{Exploration of model prompt sensitivity and boundary conditions. By comparing single point, multiple points, and bounding box interactions, this figure reveals the limitations of the model when dealing with out-of-distribution noise (left column) and ambiguous local features (middle column), using the right column as a baseline success reference.}}
  \label{fig10}
\end{figure}

{To comprehensively evaluate the boundary conditions and prompt sensitivity of the proposed MedP-CLIP framework, we conduct an in-depth case study analyzing typical failure modes alongside successful predictions. As illustrated in Fig. \ref{fig10}, we evaluate three representative clinical scenarios across single-point, multi-point, and bounding box interaction modalities to understand how prompt granularity influences model reasoning and feature decoupling.}

\subsection*{{Case1: Impact of Artifacts and Noise}}

{The first column of Fig. \ref{fig10} demonstrates the vulnerability of current VLM-based frameworks to out-of-distribution noise. In this case, severe artifacts disrupt the normal boundary and texture of the kidney, resulting in an abnormal intensity pattern resembling intestinal structures. When restricted to a highly localized "single-point" prompt, the visual encoder struggles to extract reliable structural embeddings, leading to a misclassification of the region as the small intestine (confidence score 0.309 vs. kidney 0.274). However, providing denser prompts (i.e., multi-point or bounding box) successfully rectifies this error. This indicates that broader spatial constraints introduce sufficient surrounding context, effectively counteracting local pixel degradation and highlighting the practical advantage of MedP-CLIP's flexible prompting mechanism in noisy environments.}

\subsection*{{Case2: Semantic Confusion in Similar Adjacent Structures}}

{The second column reveals a critical limitation stemming from semantic and morphological ambiguity. Liver cysts typically manifest as well-defined low-density areas, sharing highly similar visual characteristics with the adjacent gallbladder. Under a "single-point" prompt, the model relies on overly localized appearance cues and misclassifies the cyst as the gallbladder. Conversely, under a "bounding box" prompt, the model is misled because the elongated shape framed by the box strongly mimics the typical morphological appearance of the pancreas in 2D CT slices, resulting in misclassification. Accurate prediction is only achieved using the "multi-point" prompt, which provides sufficient constraints to outline the irregular lesion accurately. This failure mode underscores the challenge of relying solely on 2D visual cues without explicit 3D anatomical priors, which occasionally leads to confusion among structures with similar visual or morphological profiles.}

\subsection*{{Case3: Prompt Robustness in Distinct Anatomy}}

{For comparison, the third column presents a standard success case involving a gallbladder with exceptionally clear anatomical boundaries. MedP-CLIP exhibits strong robustness here, achieving correct predictions across all three prompt types. Notably, as the prompt's information density increases from a single point to a bounding box, the model's prediction confidence exhibits a steady upward trajectory (0.728, 0.772, and 0.833, respectively). This validates the efficacy of our feature-level prompt fusion mechanism, proving that in images with low semantic ambiguity, a bounding box can perfectly encapsulate the complete anatomical context to maximize discriminative confidence.}

{The overarching analysis reveals that the quality and granularity of interactive prompts are directly correlated with the model's localization capabilities. While empowering clinicians to flexibly adjust interaction modalities (e.g., switching from bounding boxes to multi-point prompts for blurred boundaries) significantly enhances practical usability, resolving the fundamental confusion of adjacent semantic structures remains an open challenge. In future work, integrating global 3D anatomical atlases into the spatial reasoning module will be a crucial direction to provide explicit structural priors and further mitigate prompt sensitivity.}

\subsection{{Discussion}}


While MedP-CLIP builds upon the foundational success of vision-language pre-training and promptable architectures, its core contribution lies in a feature-level region prompt integration mechanism tailored for medical vision-language representation learning. Rather than replacing CLIP with a fundamentally new architecture, MedP-CLIP extends the CLIP framework with lightweight prompt-conditioned adaptation, aiming to reduce the gap between global image-text alignment and region-specific diagnostic reasoning. To further clarify the design characteristics of MedP-CLIP and its relationship to existing methods, we systematically compare it with representative SOTA models across four dimensions.

\begin{table}[pos=htbp]
\footnotesize
\centering
\caption{Systematic comparison of MedP-CLIP with SOTA models across key architectural and functional dimensions}
\resizebox{\textwidth}{!}{
\begin{tabular}{lccccc}
\toprule[1.5pt]
\textbf{Model} & \makecell{\textbf{Medical} \\ \textbf{VLM}} & \makecell{\textbf{Feature-Level} \\ \textbf{Integration}} & \makecell{\textbf{Region-Aware} \\ \textbf{Representation}} & \makecell{\textbf{Region-Text} \\ \textbf{Pre-training}} & \makecell{\textbf{Unified} \\ \textbf{Foundation}} \\
\midrule
SAM \citep{sam} & \xmark & \cmark & \cmark & \xmark & \xmark \\
Alpha-CLIP \citep{sun2024alpha} & \xmark & \xmark & \cmark & \cmark & \cmark \\
UniMed-CLIP \citep{khattak2024unimed} & \cmark & \xmark & \xmark & \xmark & \cmark \\
ASG \citep{li2024anatomical} & \cmark & \xmark & \cmark & \cmark & \xmark \\
R-LLaVA \citep{chen2025r} & \cmark & \xmark & \cmark & \xmark & \xmark \\
MedP-CLIP & \cmark & \cmark & \cmark & \cmark & \cmark \\
\bottomrule[1.5pt]
\end{tabular}
}
\end{table}

\subsection*{{(1) Feature-level vs. Input-level Integration}}

{Promptable models have revolutionized general computer vision, yet directly adapting them to medical VLMs poses severe structural challenges. SAM exhibits profound zero-shot segmentation capabilities via prompt encoders. However, SAM operates strictly within the vision domain; it lacks a text encoder and cannot perform cross-modal semantic reasoning, rendering it incapable of tasks like diagnostic visual question answering (VQA) or region-text retrieval. To introduce region-awareness into VLMs, models like Alpha-CLIP rely on an input-level modification, fusing RoI information via an auxiliary alpha channel. This architectural choice necessitates dense masks at inference time, which is clinically impractical. In contrast, MedP-CLIP introduces a native feature-level integration mechanism. By processing points, bounding boxes, and masks into sparse and dense embeddings that interact directly with image tokens via cross-attention, MedP-CLIP decouples its reasoning pipeline from the need for pixel-level masks or auxiliary input channels. This allows for dynamic, lightweight interaction without altering the original image space.}

\subsection*{{(2) Region-Aware vs. Global Encoders}}

{Recent advancements in medical VLMs, including MedCLIP, BiomedCLIP, UniMed-CLIP, and the highly scaled BMC-CLIP (trained on 24M images), have significantly pushed the boundaries of medical image-text alignment. However, these models are fundamentally constrained by their global pooling architectures. They map entire images to holistic textual descriptions, effectively averaging out critical spatial nuances. Since clinical diagnoses often hinge on subtle, highly localized pathologies (e.g., a millimeter-sized nodule in a full chest CT), these global encoders inevitably fail at fine-grained localization. MedP-CLIP inherently bridges this gap. By conditioning the visual representation on spatial prompts, it maintains the robust global semantic alignment of baseline CLIPs (as evidenced by our superior zero-shot classification performance) while unlocking improved regional comprehension compared with global models.}

\subsection*{{(3) Region Image-Text vs. Global Image-Text}}

{To support these architectural innovations, MedP-CLIP employs the largest medical region-text dataset to date. While global models typically rely on scraping vast amounts of image-text pairs, this approach inherently lacks spatial specificity. Instead of merely increasing the number of global images, we focus on mining local semantic value, extracting 97.3 million fine-grained region-level annotations and structured clinical diagnostic descriptions from 6.4 million medical images. This significantly surpasses existing medical VLM datasets in both scale and breadth, seamlessly covering 15 different imaging modalities and six major anatomical groups. Furthermore, to ensure the network learns near-authentic medical representations, we synthesize these textual data through an ontology-based process, strictly adhering to the constraints of the RadLex Lexicon v4.1. Human sampling validation confirms a data validity rate of 98.4\%.}

\subsection*{{(4) Native vs. Task-Specific}}

{Recognizing the limitations of global encoders, several recent methods have attempted to introduce region-awareness, but typically through fragmented, task-specific pipelines. Models like ASG and CARZero achieve anatomical- or patch-level alignment during training, but they do not support explicit, user-driven spatial prompts during inference. Methods such as R-LLaVA incorporate spatial constraints by directly merging bounding box annotations into the input image space via alpha blending, alongside passing spatial coordinates into the text prompt of the LLMs. While effective for specific tasks, modifying the raw image pixels and relying on the language model to interpret spatial coordinates can disrupt the unified, end-to-end nature of the visual encoder. Unlike these fragmented approaches, MedP-CLIP is designed as a scalable, plug-and-play visual backbone. It natively supports intuitive clinician prompts (points, boxes) without requiring external feature extractors, coordinate-to-text transformations, or destructive image operations like cropping or pixel blending. This enables seamless deployment across diverse clinical applications from interactive segmentation to multimodal clinical reasoning, establishing a unified foundational capability for medical AI.}

\begin{table}[pos=htbp]
\centering
\caption{Ablation studies and comparative analysis. (a) Impact of prompt ensembling on zero-shot performance across distinct templates. (b) Effect of varying the number of unfrozen Transformer blocks. (c) Performance comparison with UniMed-CLIP under equivalent data scaling (5.3M Images).}
\label{tab:all_ablations}
\vspace{-2pt} 

\begin{subtable}[pos=htbp]{\textwidth}
\centering
\caption{The result of prompt ensembling on the COVID-CT and ACL datasets}
\label{sup_tep}
\setlength{\tabcolsep}{28pt} 
\renewcommand{\arraystretch}{1.} 
\begin{tabular}{lcc} 
\toprule[1.5pt]
 & COVID-CT & ACL \\ \hline
Best Template & 68.47\% & 59.35\%  \\
Worst Template & 63.55\%  & 49.66\%  \\
Ensembling Templates & 67.49\%  & 51.71\%  \\ 
\bottomrule[1.5pt]
\end{tabular}
\end{subtable}

\vspace{10pt}

\begin{subtable}[t]{0.35\textwidth} 
\centering
\caption{Unfreeze block ablation}
\label{sup_unfree}
\begin{tabular}{cc}
\toprule[1.5pt]
\textbf{Unfreeze Block} & \textbf{Top-1 Acc} \\
\midrule
2 & 86.54\% \\
4 & 87.24\% \\
6 & 87.96\% \\
8 & 88.64\% \\
10 & 89.41\% \\
12 & 90.24\% \\
\bottomrule[1.5pt]
\end{tabular}
\end{subtable}%
\hspace{-0.8cm} 
\begin{subtable}[t]{0.45\textwidth} 
\centering
\caption{Comparison with UniMed-CLIP}
\setlength{\tabcolsep}{4pt} 
\label{tab:clip_comparison}
\begin{tabular}{lcc}
\toprule[1.5pt]
\textbf{Dataset} & \textbf{UniMed-CLIP} & \textbf{MedP-CLIP} \\
\midrule
COVID-CT & 61.08\% & 66.70\% \\
ACL & 43.78\% & 58.63\% \\
ChestCT & 35.40\% & 32.85\% \\
ACRIMA & 59.32\% & 55.21\% \\
Br35H & 74.59\% & 90.28\% \\
Avg. Acc & 54.83\% & 60.73\% \\
\bottomrule[1.5pt]
\end{tabular}
\end{subtable}

\end{table}

\section{Ablation Studies}\label{s5}

\subsection{Test-time Prompt Ensembling}

As shown in Tab.~\ref{sup_tep}, our evaluation demonstrates that the choice of prompt templates in zero-shot inference considerably affects performance. Although certain prompts can achieve superior results on specific datasets or modalities, using a single template often leads to inconsistent outcomes due to its sensitivity to phrasing and domain alignment. In comparison, ensembling multiple prompts provides a more balanced and robust strategy. While not always attaining peak performance, this approach mitigates the instability associated with individual templates. To improve generalization and reduce performance variance, we recommend creating diverse and context-aware prompt templates adapted to the characteristics and modalities of each dataset. These can be designed manually or generated automatically using language models such as GPT~\citep{brown2020language}. Ensembling such prompts can then enhance the robustness and stability of zero-shot inference.

\begin{figure}[pos=h]
  \centering
  \includegraphics[width=0.75\linewidth]{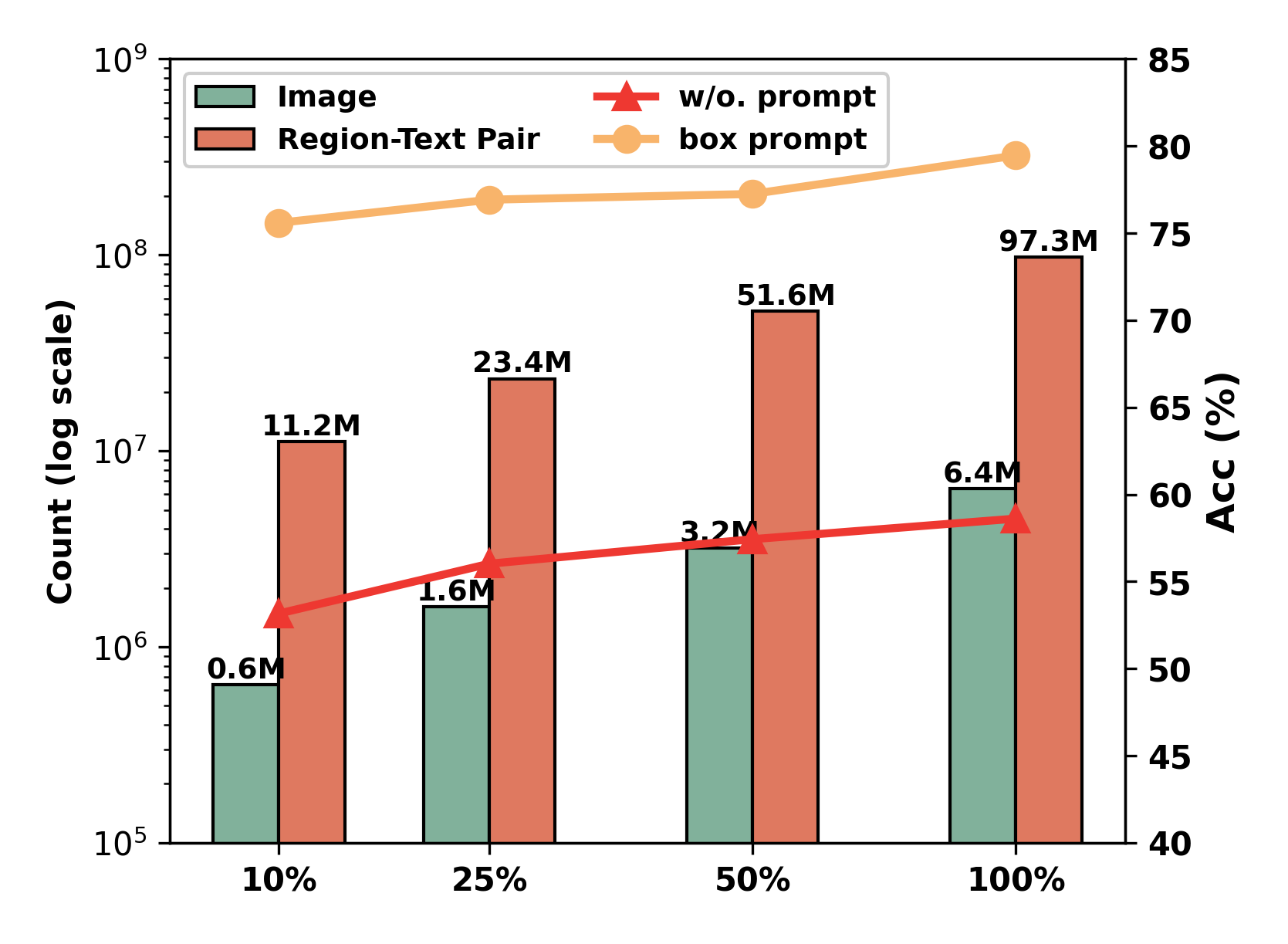}
  \caption{The relationship between MedP-CLIP pre-training data scaling and zero-shot classification.}
  \label{fig11}
\end{figure}

\subsection{Encoder Unfreezing}

Optimizing MedP-CLIP performance, we trained the model on the TotalSegmentator dataset~\citep{wasserthal2023totalsegmentator} while unfreezing varying numbers of vision encoder layers. Experiments employed the ViT-B/16 architecture~\citep{CLIP} (containing 12 Transformer blocks). Model performance was evaluated via zero-shot Top-1 accuracy for regional (anatomical structure) classification on the AMOS2022 dataset~\citep{ji2022amos}. As shown in Tab.~\ref{sup_unfree}, we incrementally increased the number of unfrozen Transformer blocks with a step size of 2 (ranging from 2 to fully unfrozen blocks). Results demonstrate that classification accuracy exhibits a consistent upward trend as the number of unfrozen layers increases.

\subsection{Data Volume Scaling}

We investigated the impact of data volume on enhancing model robustness through ablation studies. The study conducted pre-training on the ViT-L/14 model using 10\%, 25\%, 50\%, and 100\% of the dataset, followed by zero-shot classification without prompts and box-prompted classification on the Br35H dataset~\citep{wageh2024brain}. As shown in Fig.~\ref{fig11}, the number of training images increased from 0.6M to 6.4M, while the region-text pairs grew from 11.2M to 97.3M. The model's classification performance improved with larger pre-training datasets, demonstrating that data volume is crucial for model learning capability. Notably, when box prompts were used, the model's classification accuracy significantly improved. {Furthermore, to isolate our methodological advantages from the benefits of mere data scaling, we conducted a comparative analysis against UniMed-CLIP under an equivalent data volume constraint of 5.3M images. As detailed in Table~\ref{tab:all_ablations}(c), our MedP-CLIP achieves a superior average accuracy of 60.73\% across five datasets, yielding a notable 5.9\% improvement over UniMed-CLIP (54.83\%). Although UniMed-CLIP maintains a slight edge on the ChestCT and ACRIMA datasets, MedP-CLIP demonstrates substantial gains on COVID-CT, ACL, and particularly Br35H, confirming the inherent effectiveness and robust zero-shot generalizability of our proposed framework even when trained on identical data scales.}

\section{Conclusion}
MedP-CLIP is a region-aware VLM addressing the pressing need for fine-grained spatial understanding in medical imaging. By integrating medical prior knowledge and proposing a versatile feature-level region prompt integration mechanism, our model responds to various spatial prompts while preserving essential global context, mitigating limitations of previous methods like cropping or masking that disrupt context. Pre-trained on a meticulously constructed medical image region-text dataset, MedP-CLIP demonstrates state-of-the-art performance across multiple clinically relevant tasks, including image-level zero-shot recognition, region-level retrieval/classification, interactive segmentation, and multimodal large language model reasoning. It seamlessly integrates with existing CLIP-based pipelines, providing a scalable, plug-and-play visual backbone that synergistically combines holistic image understanding with precise regional analysis, significantly enhancing medical AI capabilities.

{Although MedP-CLIP extends CLIP from holistic global representation toward fine-grained, region-aware reasoning, several avenues remain for future refinement. First, while our feature-level integration natively accommodates flexible prompts and successfully decouples reasoning from dense clinical masks, it occasionally exhibits semantic confusion when differentiating morphologically similar adjacent structures within 2D slices. Addressing this morphological ambiguity necessitates the future integration of global 3D anatomical atlases to provide explicit structural priors. Second, although MedP-CLIP excels as a plug-and-play backbone in interactive, clinician-in-the-loop workflows, its current reliance on explicit spatial guidance poses challenges for fully automated, high-throughput screening environments. Consequently, empowering the model to autonomously infer implicit regions of interest directly from unstructured diagnostic narratives represents a critical next step to reduce workflow friction. This direction is also consistent with the emerging trend of agentic AI in clinical care, where autonomous perception and reasoning are increasingly regarded as important components for reducing clinical workflow burden and supporting more adaptive decision-making processes~\citep{zou2026agentic}. Finally, adapting the architecture to support higher-resolution inputs for micron-scale pathology identification will further solidify MedP-CLIP as a comprehensive and scalable foundation for next-generation medical AI.}

\appendix
\section{Region-Aware Text Generation Methodology}
\label{sup_qcm}

\subsection*{(1) GPT-4o-based Region-Aware Text Generation Templates}

To ensure reproducibility of the region-level text generation process, we provide the complete prompt protocol used for GPT-4o-based description generation. The prompt consists of a fixed system instruction, a structured user template, a constrained output schema, and automatic regeneration rules. GPT-4o was used only to convert structured image-mask-category evidence into fluent region-level descriptions, rather than to infer unobserved clinical findings.

\subsubsection*{System Prompt}

\begin{quote}
You are a medical imaging text generation assistant. Your task is to generate concise, clinically realistic descriptions for local anatomical or pathological regions in medical images. To assist your descriptions, strictly reference and use terminology from the \{Radiology Lexicon\}. You must strictly adhere to the provided image metadata, mask information, and category labels. Do not infer unobserved diagnoses, disease severity, patient attributes, treatment recommendations, or clinical outcomes. All generated descriptions must be evidence-based, anatomically valid, and suitable for vision-language pre-training.
\end{quote}

\subsubsection*{User Prompt Template}
For each image-mask-category triplet $\{I, M_k, C_k\}$, we use the following structured prompt:

\begin{quote}
Given the following medical image region information:

\textbf{Modality}: \{modality\} \\
\textbf{Anatomical group}: \{anatomical\_group\} \\
\textbf{Region category}: \{category\_label\} \\
\textbf{Mask location}: \{mask\_location\} \\
\textbf{Mask geometry}: \{mask\_shape\_descriptor\} \\
\textbf{Optional intensity or appearance cue}: \{appearance\_cue\}

Generate one region-aware medical image description. The description should:
(1) mention the imaging modality when appropriate;
(2) describe the highlighted or masked region using the given category label;
(3) refer to the spatial or anatomical context if provided;
(4) use standardized and non-speculative medical terminology;
(5) avoid any diagnosis or finding that is not explicitly supported by the provided information.

Return the result in the following JSON format:

\{ \\
``description'': ``...'', \\
``modality'': ``\{modality\}'', \\
``region\_category'': ``\{category\_label\}'' \\
\}

\end{quote}

\subsection*{{(2) Output Constraints}}

The generated description must satisfy the following constraints:
\begin{itemize}
    \item The description must be grounded in the input triplet $\{I, M_k, C_k\}$.
    \item The description must explicitly refer to the masked or highlighted region.
    \item The description must not introduce unprovided diagnoses, disease stages, patient demographics, temporal progression, treatment recommendations, or prognostic statements.
    \item Speculative expressions such as ``likely'', ``possibly'', ``suggestive of'', ``may indicate'', and ``consistent with'' are prohibited unless such information is explicitly provided by the original annotation.
    \item The output must follow the predefined JSON schema. Outputs with missing fields, invalid format, or unsupported clinical claims are discarded and regenerated.
\end{itemize}

\subsection*{{(3) Automated Quality Control}}

{To further ensure the diversity and clinical precision of the corpus, we implemented multi-stage quality control and semantic distinctness filtering strategies. During the automated verification phase, the system checks for missing core elements (e.g., modality, spatial location) and validates the image-text spatial consistency of the triplets. To prevent the model from collapsing into generic linguistic patterns, we quantify the semantic distinctness by computing the cosine similarity between the newly generated descriptions and existing instances. Specifically, given a newly generated description $t_{\text{new}}$ and an existing description $t_j$ from the established corpus $\mathcal{T}$, we extract their high-dimensional embeddings $\mathbf{v}_{\text{new}} = \Phi(t_{\text{new}})$ and $\mathbf{v}_j = \Phi(t_j)$ using a text encoder $\Phi$. The semantic similarity score $\mathcal{S}$ is formulated as:}

{$$\mathcal{S}(t_{\text{new}}, t_j) = \frac{\mathbf{v}_{\text{new}}^\top \mathbf{v}_j}{\|\mathbf{v}_{\text{new}}\|_2 \|\mathbf{v}_j\|_2}$$
If the maximum semantic overlap exceeds a predefined threshold $\tau$ (i.e., $\max_{t_j \in \mathcal{T}} \mathcal{S}(t_{\text{new}}, t_j) > \tau$, where $\tau = 0.9$), or if contradictory descriptions for the same anatomical region emerge (e.g., describing the exact same region as both "hypodense" and "isodense"), the instance is automatically discarded, triggering a regeneration mechanism. This mathematical constraint forces the data distribution to align naturally with clinical variations and explicitly prevents linguistic homogenization.}

\begin{table}[t]
\centering
\small
\caption{Sensitivity analysis of the semantic filtering threshold $\tau$ on AMOS2022.}
\label{threshold_sensitivity}
\begin{tabular}{c c c c c}
\toprule[1.5pt]
\textbf{Threshold $\tau$} 
& \textbf{Retention Ratio (\%)} 
& \textbf{Distinct-2} 
& \textbf{Clinical Validity (\%)} 
& \textbf{Top-1 Acc (\%)} \\
\midrule
0.85 & 82.6 & \textbf{42.8} & 96.7 & 76.9 \\
0.90 & 86.4 & 39.5 & \textbf{97.4} & \textbf{77.6} \\
0.95 & \textbf{93.1} & 34.7 & 96.9 & \textbf{77.6} \\
\bottomrule[1.5pt]
\end{tabular}
\end{table}

To justify the choice of the semantic filtering threshold, we evaluated three candidate thresholds on AMOS2022, with $\tau \in \{0.85, 0.90, 0.95\}$. For each setting, the model was fine-tuned only on the training split and evaluated on the held-out validation split. We compared the thresholds in terms of corpus retention ratio, lexical diversity, expert-validated clinical validity, and downstream region-level retrieval performance. As shown in Table.~\ref{threshold_sensitivity}, $\tau=0.85$ achieves the highest lexical diversity but removes more samples and slightly degrades downstream performance, likely because many clinically valid descriptions share standardized anatomical and modality-specific terminology. Conversely, $\tau=0.95$ retains more samples but introduces more semantically redundant descriptions, reducing corpus diversity and weakening region-level alignment. Although $\tau=0.90$ and $\tau=0.95$ obtain comparable Top-1 accuracy, $\tau=0.90$ provides higher lexical diversity and clinical validity, thereby achieving the best overall balance. Therefore, we adopt $\tau=0.90$ in the final data generation pipeline.

\subsection*{{(4) Manual Sampling Validation}}
{Beyond automated constraints, we introduced a rigorous double-blind expert review mechanism. A stratified random sample of 15,374 cross-modality and cross-organ descriptions was independently evaluated by board-certified radiologists. The evaluation criteria encompassed anatomical plausibility, image-semantic consistency, terminology standardization, and visual evidence fidelity. The audit revealed exceptionally high inter-rater reliability (Cohen's $\kappa=0.87$), with the generated medical texts achieving a clinical validity rate of 98.4\% and an overall rejection rate strictly maintained below 2\%. This robustly demonstrates that our generation strategy effectively captures authentic medical knowledge rather than simply fitting an AI-like stylistic template.}

\begin{figure}[pos=htbp]
  \centering
  \includegraphics[width=0.8\linewidth]{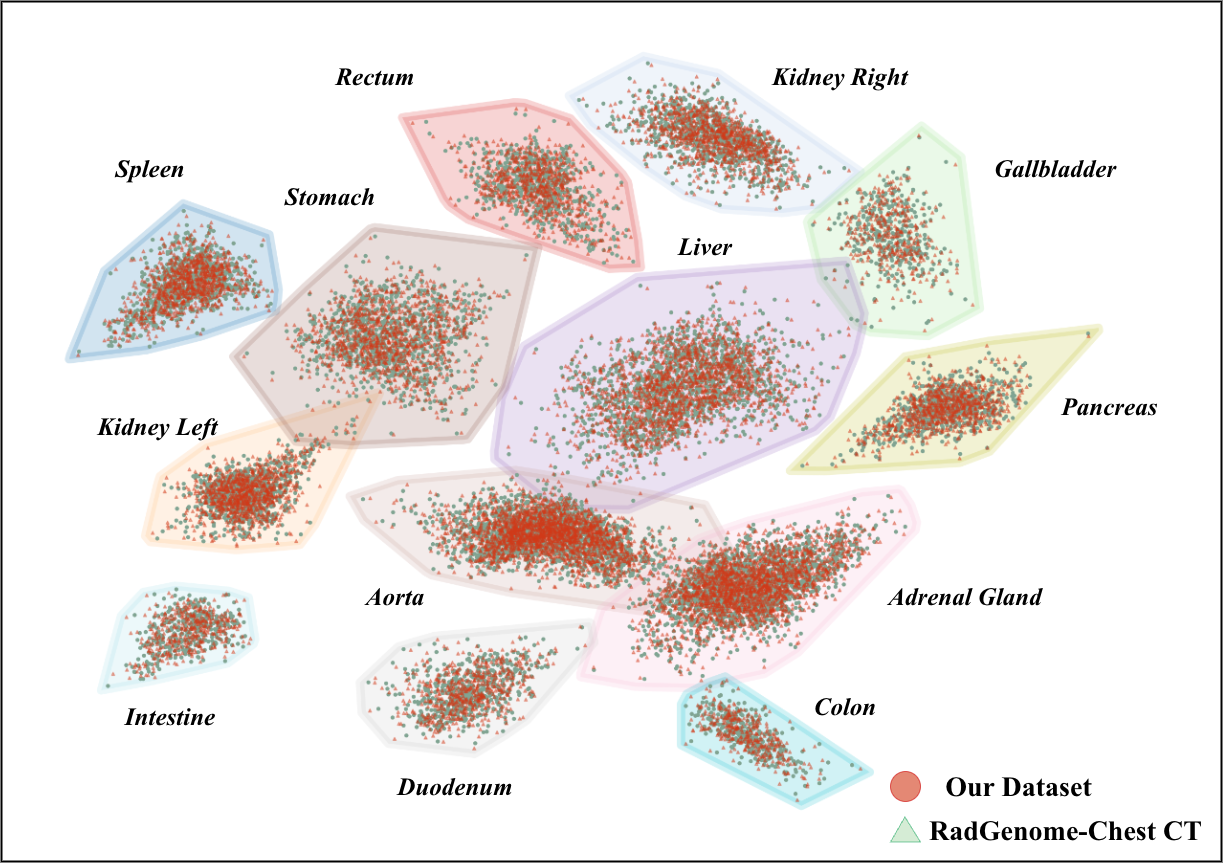}
  \caption{{t-SNE visualization of semantic embeddings from our synthesized clinical descriptions (red circles) and human-annotated RadGenome-Chest CT descriptions (green triangles) across 13 anatomical categories.}}
  \label{fig12}
\end{figure}

\subsection*{{(5) Semantic Feature Distribution Validation}}

{To provide intuitive empirical support from the feature space perspective, we conducted a quantitative analysis of the semantic and linguistic distribution of the synthesized texts (as illustrated in Fig. \ref{fig12}). We randomly sampled descriptions corresponding to 13 distinct abdominal organ categories from both our synthetic dataset and the expert-annotated RadGenome-Chest CT dataset. We extracted semantic embeddings using a frozen domain-specific text encoder (BioClinicalBERT) and applied t-SNE for dimensionality reduction. The results demonstrate that the synthesized descriptions (red circles) and human expert annotations (green triangles) exhibit near-complete overlap and co-distribution within every anatomical cluster. This visualization reveals two critical characteristics: First, high intra-class alignment proves that within the same organ category, the synthetic data seamlessly integrates into the human-annotated data distribution, rather than isolating into an artificial GPT linguistic space. Second, strong inter-class separability indicates that the generated texts accurately preserve highly specific, organ-level medical features without collapsing into ambiguous, generic patterns.}


\bibliographystyle{cas-model2-names}

\bibliography{cas-refs}

\end{document}